\documentclass{bmvc2k}

\usepackage{amsfonts}       

\usepackage{paralist}
\usepackage{multirow}
\usepackage[normalem]{ulem}

\def\eg{\emph{e.g.},} 
\def\ie{\emph{i.e.},}

\newcommand{\Eref}[1]{Eq.~(\ref{#1})}
\newcommand{\Fref}[1]{Fig.~\ref{#1}}

\newcommand{\Cref}[1]{Chap.~\ref{#1}}
\newcommand{\Sref}[1]{Sec.~\ref{#1}}

\title{Single-Modal Entropy based Active Learning for Visual Question Answering}

\addauthor{Dong-Jin  Kim*}{djnjusa@gmail.com}{0}
\addauthor{Jae  Won  Cho*}{chojw@kaist.ac.kr}{0}
\addauthor{Jinsoo Choi}{jinsc37@kaist.ac.kr}{0}
\addauthor{Yunjae  Jung}{yun9298a@gmail.com}{0}
\addauthor{In  So  Kweon}{iskweon77@kaist.ac.kr}{0}

\addinstitution{
 Korea Advanced Institute of Science
and Technology (KAIST)\\
 Daejeon, South Korea\\
 (* indicates equal contribution)
}

\def\eg{\emph{e.g}\bmvaOneDot}

\runninghead{Kim~et al.}{Single-Modal Entropy based Active Learning}

\begin{document}

\maketitle

\begin{abstract}
Constructing a large-scale labeled dataset in the real world, especially for high-level tasks (\eg,~Visual Question Answering), can be expensive and time-consuming. In addition, with the ever-growing amounts of data and architecture complexity, Active Learning has become an important aspect of computer vision research.
In this work, we address Active Learning 
in the multi-modal setting of Visual Question Answering (VQA).
In light of the multi-modal inputs, image and question, we propose a novel method for effective sample acquisition through the use of ad hoc single-modal branches for each input to leverage its information. 
Our mutual information based sample acquisition strategy Single-Modal Entropic Measure (SMEM)
in addition to our self-distillation technique enables the sample acquisitor to exploit all present modalities and find the most informative samples.
Our novel idea is simple to implement, cost-efficient, and 
readily adaptable to other multi-modal tasks.
We confirm our findings on various VQA datasets through state-of-the-art performance by comparing to existing Active Learning baselines.
\end{abstract}

\section{Introduction}


Recent successes of learning-based computer vision methods rely heavily on abundant annotated training examples, which may be prohibitively costly to label or impossible to obtain at large scale~\cite{deng2009imagenet}. 
In order 
to mitigate this drawback, Active Learning (AL)~\cite{cohn1996active} has been introduced to minimize the number of expensive labels for the supervised training of deep neural networks by selecting (or \emph{sampling}) a subset of informative samples from a large unlabeled collection of data~\cite{settles2009active,tong2001support}.
AL has been shown to be a potential solution~\cite{tong2001support}
for vision~\&~language tasks 
like image captioning~\cite{shen2019learning} that rely extensively on large, expensive, curated datasets~\cite{antol2015vqa,gurari2018vizwiz,krishna2017visual,zhu2016visual7w}. 
Nonetheless, we find through experimentation that existing state-of-the-art AL strategies to these tasks result in a similar performance to each other without much distinction and without significant improvements when compared to random sampling
~\cite{figueroa2012active,mairesse2010phrase,settles2008active}. 
A key distinction in
vision~\&~language tasks is the presence of \emph{multi-modal} inputs.
Previous AL works 
have focused more on single-modal inputs; hence, studies of muli-modal AL is less prevalent. 
As multi-modal tasks also suffer the same data issues, we turn our focus to the Visual Question Answering (VQA) task as it provides a simple multi-modal input, single output classification framework.

As studies have shown that VQA models can be reliant on single modalities~\cite{goyal2017making,cadene2019rubi}, we define a novel pool-based AL strategy for VQA by leveraging mutual information of the multiple inputs, image and text, individually through a \emph{multi-branch} model. 
In this multi-branch model, we denote a ``main'' branch that relies on all modalities.
If a single-modal prediction $P(A|V)$ or $P(A|Q)$ is different from the ``main'' prediction $P(A|V,Q)$, it may signify that the missing modality is informative, where $A$, $V$, and $Q$ are the answer output, visual input, and question input respectively.
Hence, we propose a method called Single-Modal Entropic Measure (SMEM) that selects instances for labeling from the unlabeled data pool based on the differences among the predictions from the single-modal and ``main'' predictions.
We also show that we can take the differences among the predictions from the single-modal and ``main'' predictions into account by simply computing the \emph{entropy of the single-modal} predictions.
In effect, we propose an uncertainty based sampling paradigm that enables simple inference without any ad hoc steps to measure uncertainty by relying on multi-modal uncertainty.
In addition, we 
use self-distillation,
\emph{not as a regularizer}, but to directly aid in sample acquisition by forcing the single-modal branches to generate outputs similar to the ``main'' branch.
If the discrepancy between the single- and multi-modal outputs of a sample are still high despite self-distillation, that sample is highly likely to have large information gain.

Through our experimentation, we show the effectiveness of SMEM on multiple VQA datasets: VQA v2~\cite{antol2015vqa}, VizWiz VQA~\cite{gurari2018vizwiz}, and VQA-CP2~\cite{goyal2017making}.
In addition, although we design our model with multi-modal inputs in mind (\emph{e.g.}, vision-language tasks), we also evaluate our model on NTU RGB+D dataset~\cite{shahroudy2016ntu} in order to show the extensibility of our approach.
Our solution achieves favorable performance in VQA metrics~\cite{antol2015vqa} compared to existing sampling strategies~\cite{culotta2005reducing,scheffer2001active,shen2017deep,yoo2019learning,sinha2019variational,zhang2020state}, while being cost-efficient.
In particular, the proposed method is able to decrease labeling efforts by about 30\% on VQA-v2.

Our main contributions are summarized as follows:
(1) {We propose a novel sampling method, that we call Single-Modal Entropic Measure (SMEM), based on mutual information for Active Learning for VQA.}
(2) {To better sample from the single-modal branches,
we employ self-distillation between the main branch and ad hoc single-modal branches
to better train the single-modal branches for sampling.}
(3) {Through extensive experimentation, we show the effectiveness of our method on
multiple VQA datasets
compared to various competing Active Learning methods.}
(4) {We show that our method is task and architecture agnostic by testing our method on the NTU RGB+D action recognition task~\cite{shahroudy2016ntu}.}

\section{Related Work}


\noindent\textbf{Active Learning.} 
Knowing that labeling data is costly and time consuming, the task is to minimize the number of samples to annotate from a set of unlabeled data  while maximizing performance on a given task~\cite{tong2001support}. 
AL has been widely studied in image recognition~\cite{cho2021mcdal,joshi2009multi,sener2017active,yoo2019learning}, information extraction~\cite{culotta2005reducing,jones2003active,scheffer2001active}, and text categorization~\cite{hoi2006batch,lewis1994sequential}, using uncertainty-based sampling~\cite{collins2008towards,joshi2009multi}, information gain~\cite{houlsby2011bayesian}, or theoretical dropout-based frameworks~\cite{gal2017deep,kendall2017uncertainties,lin2017activelearningvqa}. Recent works leverage latent space representations using VAEs and adversarial training for improved sampling~\cite{sinha2019variational,zhang2020state,shuo2020daal}.
We do not compare to methods that require multiple steps such as Monte-Carlo dropout or model ensemble (Query-by-Committee) as they are hard to apply to recent large models for higher-level tasks. We also do not compare with the Core-Set Approach~\cite{sener2017active} as this method's computational time is directly proportional to the number of classes and the VQA v2 dataset over 3,000 classes, making this difficult to recreate. In addition, recent approaches such as~\cite{yoo2019learning,sinha2019variational,zhang2020state} have shown to outperform this approach, so we do not compare this method in our experiments.
In addition, AL has been explored in vision~\&~language tasks with varying degrees of success~\cite{chi2019justasknavigation,lin2017activelearningvqa,misra2018learning,shen2019learning}.
Although various semi-supervised learning methods for vision~\&~language tasks have been explored~\cite{kim2019image,liu2018show}, active learning has been relatively less explored.

\noindent\textbf{Visual Question Answering.}
VQA has received a considerable amount of attention due to its real-world applicability through development of several benchmark datasets~\cite{goyal2017making,johnson2017clevr,krishna2017visual,zhu2016visual7w,gurari2018vizwiz} and state-of-the-art models~\cite{fukui2016multimodal,lu2016hierarchical,yang2016stacked,kim2018bilinear}. 
Works such as~\cite{teney2018visual} and~\cite{misra2018learning} have explored VQA in a setting similar,
however~\cite{misra2018learning} achieves performance improvements by augmenting data with an additional network and~\cite{teney2018visual} focuses on meta learning tasks such as few-shot and zero-shot learning with VQA.
\cite{lin2017activelearningvqa} explores AL using existing dropout-based method without considering the multi-modality; thus, we do not experimentally compare their work.
To the best of our knowledge, we are one of the first to develop an efficient method of leveraging individual modalities to measure uncertainty for AL in VQA. 



\begin{figure}[t]
    \centering
    \includegraphics[width=.93\linewidth,keepaspectratio]{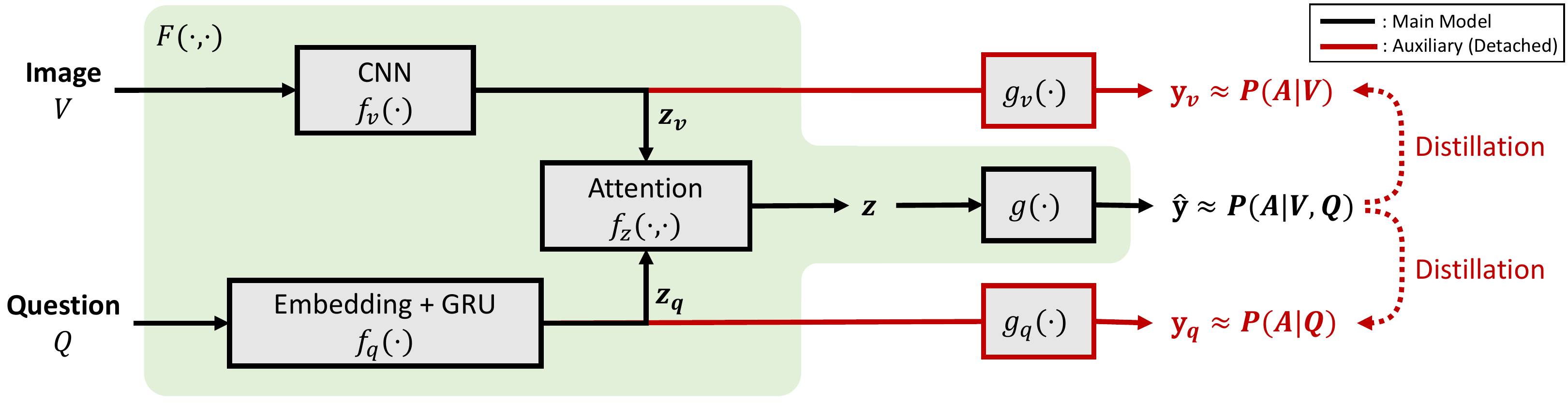}
    \caption{Illustration of our model architecture. Our model can be applied to any off-the-shelf VQA models (area in green) 
    as our goal is to design ad hoc branches solely for the purpose of \emph{measuring uncertainty}. We add two auxiliary branches $g_v(\cdot)$ and $g_q(\cdot)$ that process the visual or question features exclusively to generate answers. Along with the cross-entropy loss from ground truth answers, the single-modal branches are trained with knowledge distillation loss (\Eref{eqn:loss_q}) to follow the output from the multi-modal branch $g(\cdot)$.
    }
    \label{fig:architecture}
\end{figure}

\section{Proposed Method}

Single-Modal Entropic Measure (SMEM) consists of two key components: (1) devising an efficient and effective sample acquisition function which takes the mutual information of the individual modalities into account. (2) introducing a novel usage for the self-distillation technique to effectively train the
auxiliary branches for more effective sampling.


\subsection{Visual Question Answering Baseline}

Given an image $V$ and question $Q$ pair, a VQA task tries to correctly predict an answer from the answer set $A\in\mathcal{A}$ (in a vectorized form, $\hat{\mathbf{y}}=F(V,Q)\in [0,1]^{|\mathcal{A}|}$).
We adapt one of the famous VQA state-of-the-art model as our base architecture~\cite{anderson2018bottom} where $F(\cdot,\cdot)$ is implemented via a combination of deep neural networks.
Images and questions are mapped to feature vectors through convolutional ($f_v(\cdot)$) and recurrent ($f_q(\cdot)$) neural networks respectively (\emph{i.e.}, $\mathbf{z}_v = f_v(V), \mathbf{z}_q = f_q(Q)$).
Then following~\cite{anderson2018bottom}, $\mathbf{z}_v$ and $\mathbf{z}_q$ are combined via an additional sub-network $f_z(\cdot,\cdot)$ which uses attention to generate joint representation (\emph{i.e.}, $\mathbf{z}=f_z(\mathbf{z}_v,\mathbf{z}_q)$).
Finally, $\mathbf{z}$ is fed through a fully-connected layer $g(\cdot)$ to predict $\hat{\mathbf{y}}$ (\emph{i.e.}, $\hat{\mathbf{y}}=g(\mathbf{z})$).
The goal for training the VQA model is to follow the conditional distribution of the target dataset, $\hat{\mathbf{y}}=F(V,Q) \approx P(A|V,Q)$,
where $F(\cdot,\cdot)$ is the VQA model, and $P(\cdot)$ is the probability of the answer $A$ given image $V$ and question $Q$ 
that we want $F(\cdot,\cdot)$ to follow.
Thus, the model is trained with Binary Cross-Entropy loss compared with ground truth $\mathbf{y}$, $\mathcal{L}_{main} = BCE(\mathbf{y},\hat{\mathbf{y}})$.

\subsection{Active Learning with SMEM}
\label{sec:method}
The goal of AL is to find the best sample acquisition function $S(V,Q)$ that assigns a high score to \emph{informative} samples.
In the AL setup, a labeled set $\mathcal{D}\textsubscript{L}$ is used to train a model, then after the model converges (end of a \emph{stage}), the model is used to sample from the unlabeled set $\mathcal{D}\textsubscript{U}$, then the chosen samples are given labels and added into the labeled set $\mathcal{D}\textsubscript{L}$ and removed from the unlabeled set $\mathcal{D}\textsubscript{U}$. This step is repeated several times depending on the number of stages.
There are several ways to define the informativeness of a sample. For example, entropy defined as $H(\hat{Y})=-\sum_{i} \hat{\mathbf{y}}^i\log \hat{\mathbf{y}}^i$, defines uncertainty as informativeness.
However, in this paper, we define a novel acquisition function that takes \emph{mutual information} into account.
Our approach exploits the multi-modality of this task and directly measures the mutual information $I(A;V|Q)$ and $I(A;Q|V)$ individually 
to aid the sampling criteria. 

Intuitively, if the value of mutual information, for example $I(A;V|Q)$, is high, the variable $V$ plays an important role in predicting the answer $A$.
In order to compute mutual information of individual modalities, we
add auxiliary 
classification branches, as shown in \Fref{fig:architecture}, that predict the answer exclusively from individual visual and question features similar to~\cite{cho2021mcdal,shin2021labor}.
The branches are detached from the model 
and only utilized
to ``sample'' the unlabeled data instead of training the model; thus, the \emph{presence of the branches have no effect on the main model performance} and \emph{does not act as a regularizer} of any kind.
The auxiliary branches are also trained with BCE loss so that the output answer probability can approximate the conditional distribution of answers given cues such that $\mathbf{y}_v=g_v(\mathbf{z}_v) \approx P(A|V)$ and $\mathbf{y}_q=g_q(\mathbf{z}_q) \approx P(A|Q)$. Note however that the answer predictions from the single-modal classifiers do not have to be accurate as their purpose is only to measure uncertainty.
As we try to measure the informativeness of a given input visual question pair $(V,Q)$ through a given acquisition function $S(V,Q)$, we can intuitively expect that the input cue $V$ or $Q$ may contain a considerable amount of information if the single-modal outputs $\mathbf{y}_v$ and $\mathbf{y}_q$ are considerably different from $\hat{\mathbf{y}}$. 
Here, we devise a novel acquisition function based on entropy that takes into account the relation between the single-modal and multi-modal answer representations.

\begin{figure}[t]
    \centering
    \includegraphics[width=.9\linewidth,keepaspectratio]{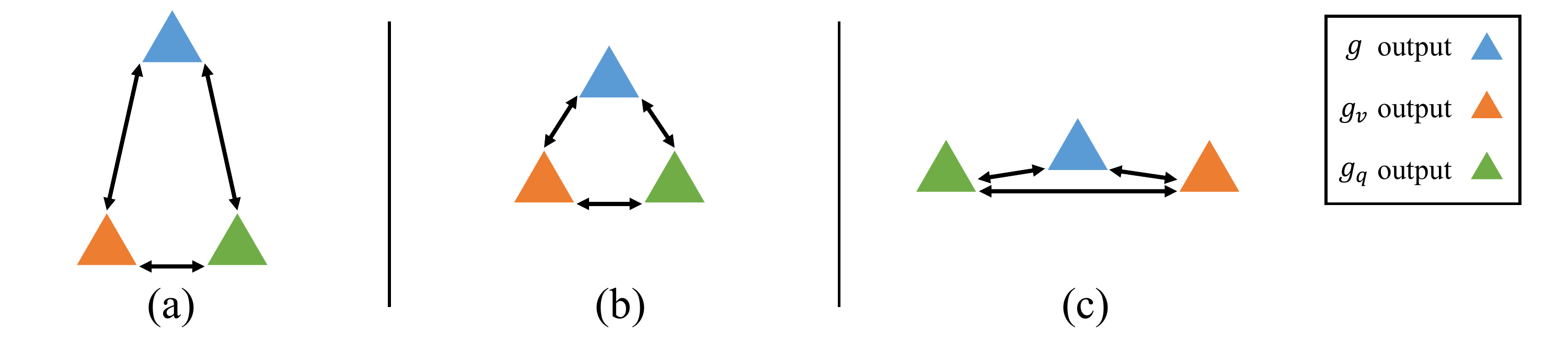}
    \caption{
    Simple illustration of the $\mathbf{y}_v$, $\mathbf{y}_v$, and $\mathbf{y}_q$.
    SMEM gives a higher score to (a) than (b) in the sampling stage. 
    Also, since we can intuitively expect (c) to be more informative than (b) when labeled, we propose to add Jensen-Shannon divergence between the two single-modal outputs in order to account for the discrepancies between single-modal outputs. 
    }
    \label{fig:jensenshanon}
\end{figure}





\noindent\textbf{Single-Modal Entropic Measure (SMEM).} 
As our final goal is to find the most informative samples, one of our objectives is to find the training data points that have high mutual information 
$I(A;V|Q)$ and $I(A;Q|V)$. According to the property of mutual information, it can be decomposed into the difference of entropies:
\begin{equation}
    I(A;V|Q) = H(A|Q) - H(A|V,Q) \approx H(Y_q) - H(\hat{Y}),
    \label{eqn:MI_Q}
\end{equation}
\begin{equation}
    I(A;Q|V) = H(A|V) - H(A|V,Q) \approx H(Y_v) - H(\hat{Y}),
    \label{eqn:MI_V}
\end{equation}
where $H(\hat{Y})=-\sum_{i} \hat{\mathbf{y}}^i\log \hat{\mathbf{y}}^i$ 
is the entropy of a given distribution. 
As shown in the equations above, we see that mutual information is easily computed from the entropy values of the single-modal and multi-modal outputs.
This is from the assumption that $H(A|V,Q)$, $H(A|Q)$, and $H(A|V)$ are modeled with $H(\hat{Y})$, $H(Y_q)$, and $H(Y_v)$ respectively (\ie~$H(A|V,Q) \approx H(\hat{Y})$, $H(A|Q)\approx H(Y_q)$, $ H(A|V)\approx H(Y_v)$).
Similarly,
rearranging \Eref{eqn:MI_Q} and \Eref{eqn:MI_V}, we get:
\begin{equation}
    H(Y_q) \approx 
    I(A;V|Q) + H(\hat{Y}),
    \label{eqn:single-Q}
\end{equation}
\begin{equation}
    H(Y_v) \approx 
    I(A;Q|V) + H(\hat{Y}),
    \label{eqn:single-V}
\end{equation}
where $H(Y_q)=-\sum_{i} \mathbf{y}_q^i\log \mathbf{y}_q^i$  and $H(Y_v)=-\sum_{i} \mathbf{y}_v^i\log \mathbf{y}_v^i$ . From \Eref{eqn:single-Q} and \Eref{eqn:single-V}, we can see that the entropy of the single-modal output is the sum of the mutual information and the entropy of the multi-modal output or "main" entropy (that we call \emph{main entropy} from here on out to avoid confusion), and we want both values to be high when sampling.
In other words,
both uncertainty and informativeness can be measured by simply computing the \emph{entropy of the single-modal output}.
The empirical advantage of leveraging the single-modal entropy instead of only using the mutual information is shown in the supplementary material.

In this regard, we propose an informativeness value to be the weighted sum of the single-modal entropies for image and question: 
\begin{equation}
    S(V,Q) = \alpha H(Y_q) + (1-\alpha) H(Y_v), 
    \label{eqn:smem}
\end{equation}
where $0\leq \alpha \leq 1$  is a hyper-parameter that weights the relative importance of visual and question scores.
Note that $0 \leq S(V,Q)\leq \log |\mathcal{A}|$, because $0 \leq H(\cdot) \leq \log |\mathcal{A}|$.

In addition, we also conjecture that as the information of the
relationships between all single- and multi-modal outputs are important, the relationship between the single-modal outputs is just as important.
As in \Fref{fig:jensenshanon} (b) and (c), even though the distance between single-modal outputs and the multi-modal output is fixed, the relationship between the three distributions can differ due to the distance between single-modal outputs.
If we only leverage the distance between the single-modal and multi-modal outputs, (b) and (c) would give the same score, however, we can intuitively expect that (c) is more uncertain than (b) as the distance between the single-modal outputs is much greater.
In order to leverage this, we also include \textbf{Jensen-Shannon divergence} (JSD) as an additional sampling criterion:
\begin{equation}
    JSD(\mathbf{y}_v||\mathbf{y}_q) = (D_{KL}(\mathbf{y}_v||M)+D_{KL}(\mathbf{y}_q||M))/2, 
    \label{eqn:jsd}
\end{equation}
where  $M = (\mathbf{y}_v+\mathbf{y}_q)/2$. 
The addition of JSD in the sampling criterion forces the sampler to find samples with a large divergence between single-modal outputs, ultimately increasing the maximum distance among all three points.
Therefore, the final function of \textbf{SMEM}:
\begin{equation}
    S(V,Q) = \alpha H(Y_q) + (1-\alpha) H(Y_v) + \beta JSD(\mathbf{y}_v||\mathbf{y}_q),
    \label{eqn:addJSD}
\end{equation}
where we introduce an additional weight hyper-parameter $\beta \geq0$.
The introduction and comparison with other possible approaches, including pure mutual information
or distance between single-modal and multi-modal branches, can be found in the supplementary material. 
We empirically determine the hyper-parameters via extensive greed search.
We also empirically found that adding the main entropy further improved the performance: $S(V,Q)+ \gamma H(\hat{Y}),$
where $\gamma \geq0$ is the hyper-parameter that weights between $S(V,Q)$ and the main entropy.
Note that $S(V,Q)+ \gamma H(\hat{Y})\geq0$, as $S(V,Q)$ and $H(\hat{Y})$ are always non-negative.
\subsection{Training Auxiliary Branches with Self-Distillation}
\label{sec:distillation}


\Eref{eqn:single-Q} and \Eref{eqn:single-V} are set on the basis that the single-modal classifiers' output distribution will be similar to that of the multi-modal classifier's especially if the visual input is conditionally independent on answers given a question (\emph{i.e.}, $P(A|VQ)\approx P(A|Q)$) or vice versa (\emph{i.e.}, $P(A|VQ)\approx P(A|V)$).
In order to satisfy this assumption, 
$g_v(\cdot)$ and $g_q(\cdot)$'s outputs distribution should be trained to resemble $g(\cdot)$'s output distribution as closely as possible.
To do so, we additionally utilize the self-distillation technique to explicitly train the auxiliary single-modal classifiers to mimic the ``main'' classifier's distribution~\cite{phuong2019distillation,zhang2019your}. Note that as the single-modal classifiers are detached, this self-distillation technique has no effect on the performance of the model and \emph{does not act as a regularizer}. Also they are \emph{not used to for any predictions}, hence having no effect on the performance as well.
Our insight is, our acquisition function $S(V,Q)$ tries to find the sample that has a large difference between the multi-modal output and the single-modal output, and the knowledge distillation loss tries to directly minimize this difference. 
In other words, if a sample gives a high $S(V,Q)$ value even after the explicit minimization of the $S(V,Q)$ score meaning the single-modal classifiers have difficulty following the output distribution of the multi-modal classifier, the sample can be thought of as a sample that is difficult or affected by both modalities.

Formally, Knowledge Distillation~\cite{hinton2015distilling} 
is to train a ``student'' (in our case, $\mathbf{y}_v$ and $\mathbf{y}_q$) to follow a ``teacher'' model or representation (in our case, $\hat{\mathbf{y}}$), which has been utilized for various proposes~\cite{cho2021dealing,hoffman2016learning,kim2018disjoint,kim2020detecting,kim2021acp++,li2016learning}. 
To this end, we add the self-distillation loss in addition to the classification loss in training $g_v(\cdot)$ and $g_q(\cdot)$ to improve our acquisition function directly. 
The auxiliary single-modal branches are trained with the following loss functions with weight hyper-parameter $\lambda \geq0$:
\begin{equation}
    \mathcal{L}_v =  BCE(\mathbf{y},\mathbf{y}_v)+\lambda BCE(\hat{\mathbf{y}},\mathbf{y}_v), \quad
    \mathcal{L}_q = BCE(\mathbf{y},\mathbf{y}_q)+\lambda BCE(\hat{\mathbf{y}},\mathbf{y}_q),
    \label{eqn:loss_q}
\end{equation}
\section{Experiments}
In this section, we describe the experimental setups, competing methods, and provide performance evaluations of the proposed methods.

\subsection{Experiments on VQA v2 Dataset}
\noindent\textbf{Dataset and metric.}
We evaluate our method on VQA v2 dataset~\cite{antol2015vqa}, which consists of 1.1M data points (visual-question-answer triplets), with 444K/214K/448K for train/val/test respectively.
Each image has at least 3 questions and 10 answers per question, thus contains 204K images, 1.1M questions, 11.1M answers, and we utilize the VQA accuracy metric~\cite{antol2015vqa}.

\noindent\textbf{Active Learning setup.}
For AL, labeled set $\mathcal{D}\textsubscript{L}$ with a size of 40,000 out of 443,757 training data points, roughly 10\% of the full dataset, is initialized randomly, and our model is trained until the training loss converges.
After converging, the end of the first stage, we use this model to collect samples from the unlabeled dataset $\mathcal{D}\textsubscript{U}$ according to an acquisition function $S(V,Q)$.
For every stage, we sample the next 40,000 data points from the unlabeled dataset and add them to the labeled dataset.
After increasing the labeled set, we train a newly initialized model using the resulting labeled set, and repeat the previous steps (marking the subsequent stages), increasing the dataset and training models for 10 iterations, resulting in a final $\mathcal{D}\textsubscript{L}$ size of 400,000 data points.

\begin{figure}[t]
\centering
\vspace{0mm}
{\includegraphics[width=.5\linewidth,keepaspectratio]{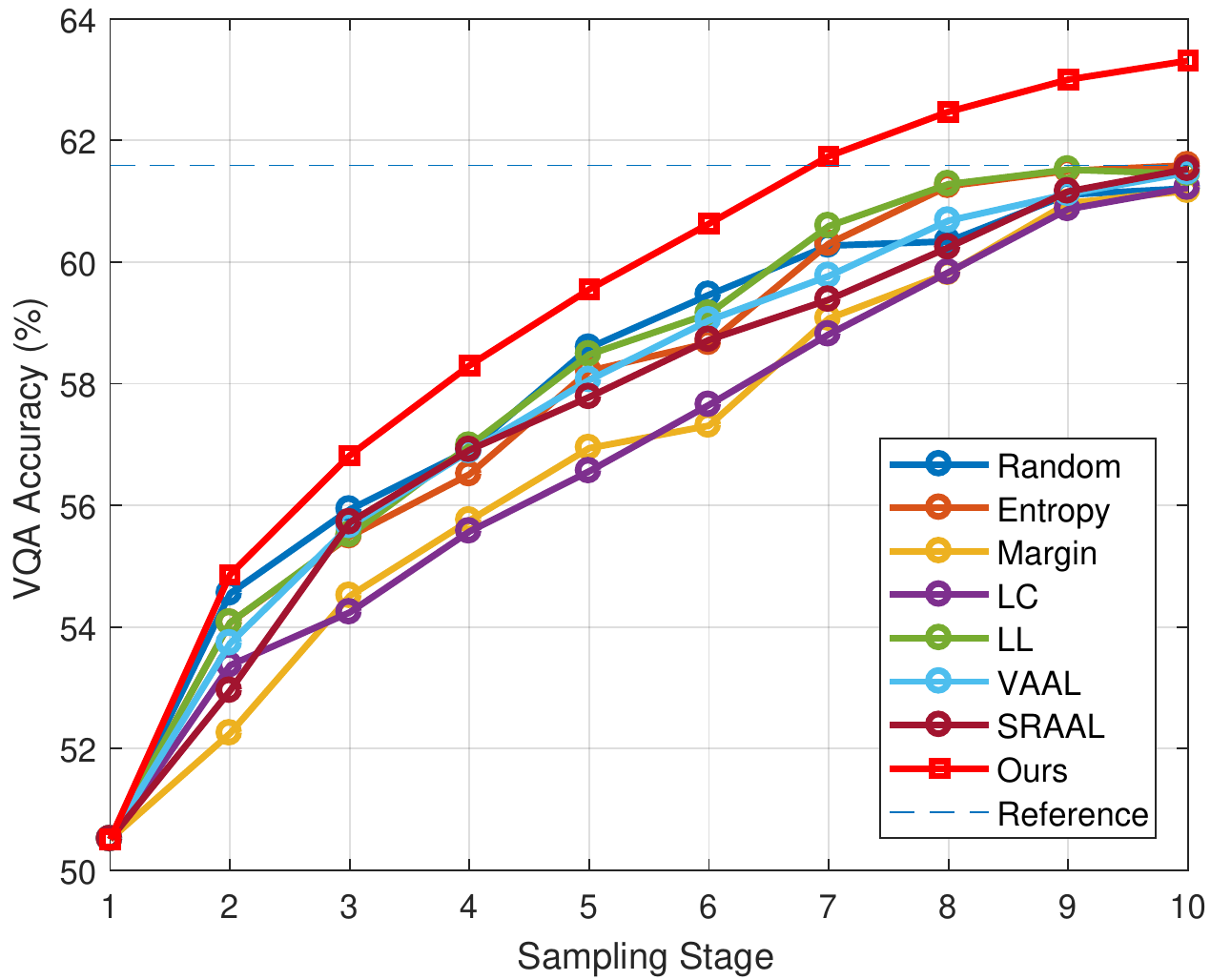}}
\caption{Comparison with existing approaches on the VQA v2 dataset. Reference is the best performance of the existing methods.
Our model shows the best performance among all the other methods by a large margin. Full dataset training at 440K samples shows 63.52\% accuracy, while ours at \emph{stage 10} at 400K samples shows 63.31\% accuracy.} 
\label{fig:quantitative}
\end{figure}


\noindent\textbf{Baseline approaches.}
In order to demonstrate the effectiveness of our method, we compare the performance of our model to widely used sample acquisition approaches~\cite{culotta2005reducing,scheffer2001active,shen2017deep} as well as the recent deep AL methods~\cite{sinha2019variational,yoo2019learning,zhang2020state}. However, we do not compare to ensemble based methods such as BALD, MC-dropout as they are inefficient and show performance worse than current state-of-the-arts such as VAAL or SRAAL.
We also do not compare to Core-Set Approach~\cite{sener2017active} as the computation cost is directly correlated to the class number, which VQA v2 has a class number of over 3,000 and they show worse performance than current state-of-the-arts such as LL4AL, VAAL, or SRAAL.
Our comparison baselines are Random (which is a uniform random distribution, sometimes called passive learning), Entropy~\cite{shen2017deep} (which is the same as ~\emph{main entropy} from \Sref{sec:method}), Margin~\cite{scheffer2001active}, Least Confident (LC)~\cite{culotta2005reducing}, LL~\cite{yoo2019learning}, VAAL~\cite{sinha2019variational}, and SRAAL~\cite{zhang2020state}. For VAAL and SRAAL, as our inputs are multi-modal, we use a joint representation for the Variational Auto-Encoder~\cite{kingma2013auto}.

\begin{figure}[t]
\centering
\begin{tabular}[c]{c}
    \subfigure[]{\includegraphics[width=.5\linewidth,keepaspectratio]{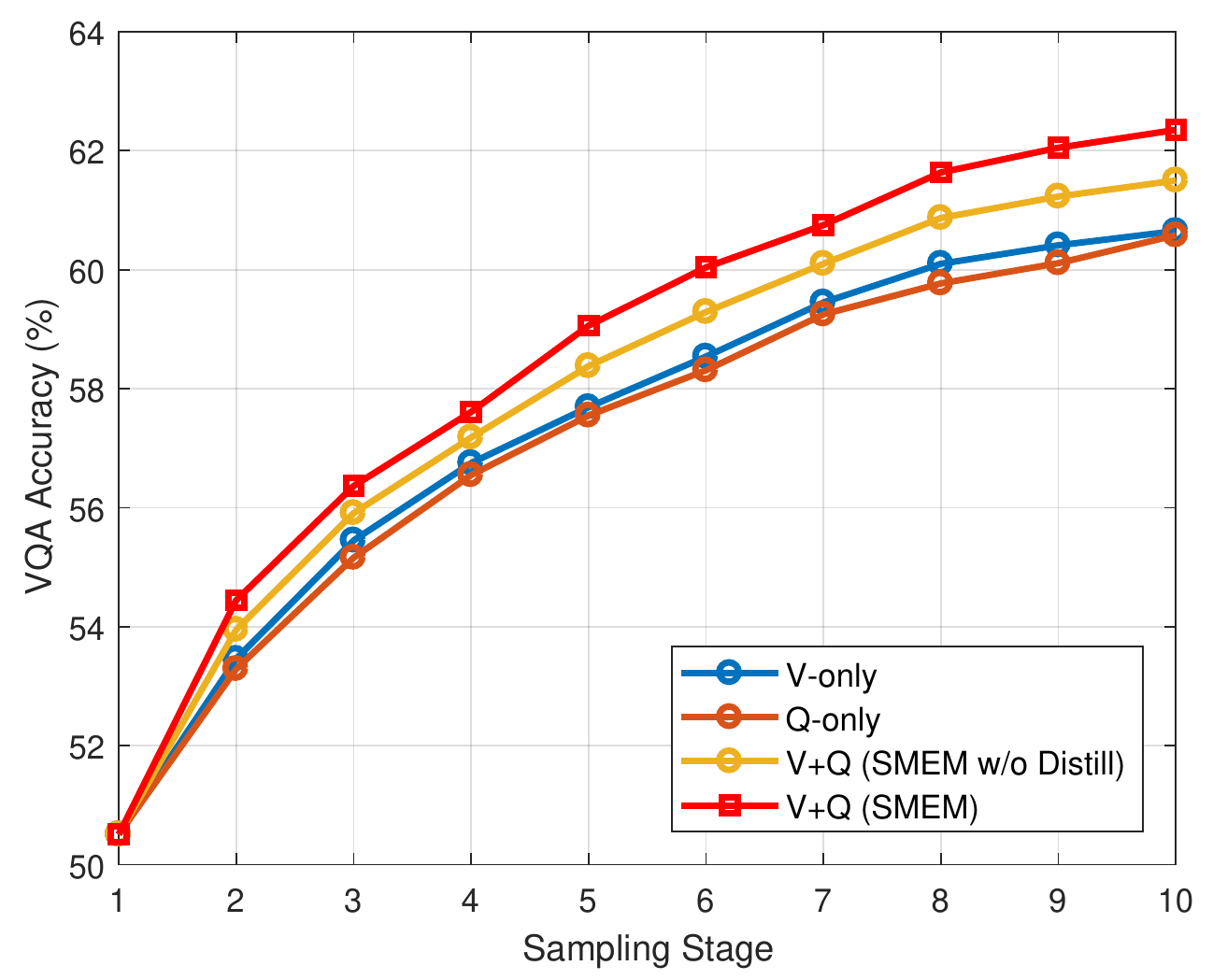}}				
    \subfigure[]{\includegraphics[width=.5\linewidth,keepaspectratio]{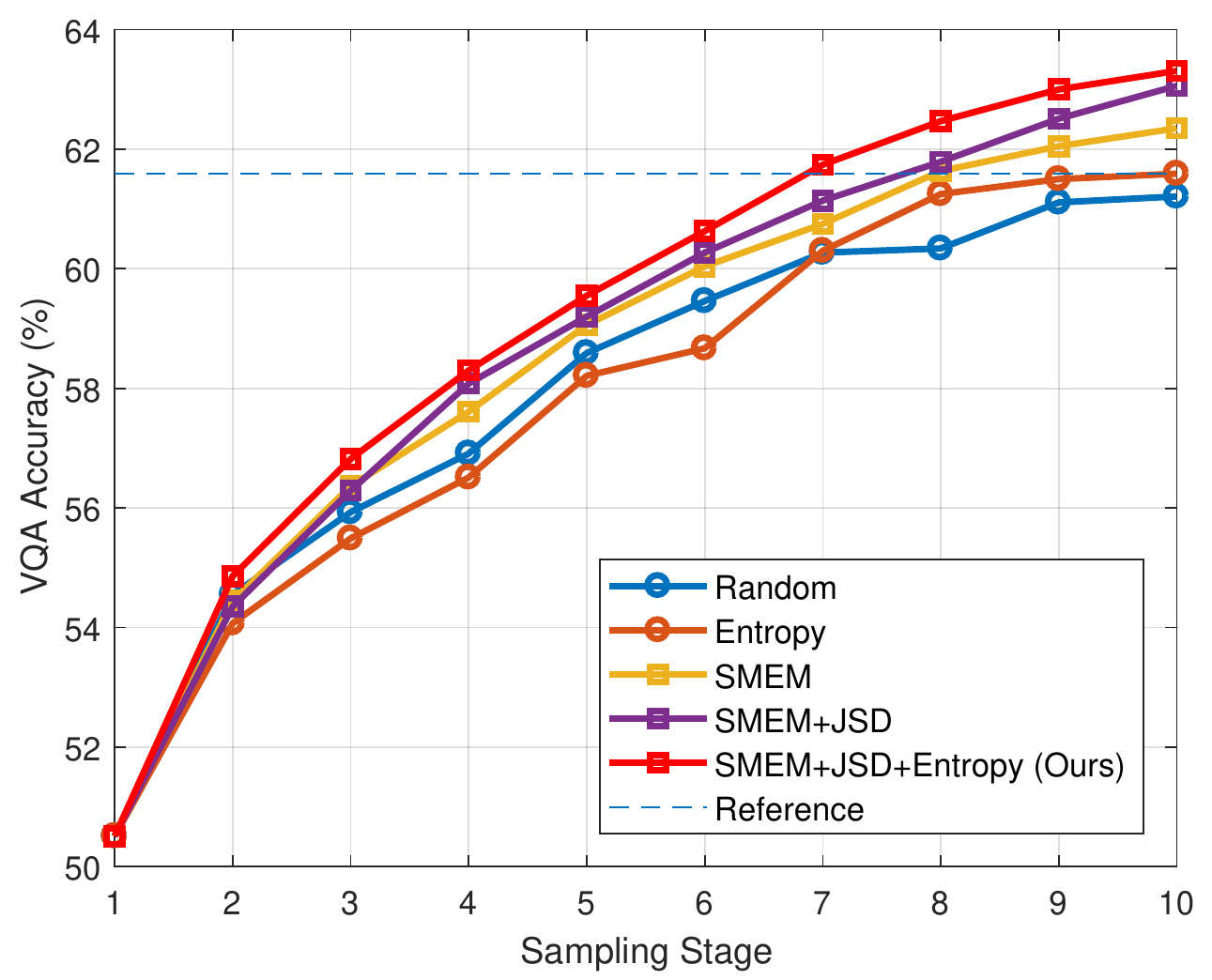}}
\end{tabular}
\caption{Ablation study of our method. 
(a) Using both V and Q modalities along with distillation (SMEM) shows the best performance in all the Active Learning stages.
(b) SMEM already shows favorable performance against traditional Active Learning methods, Random and Entropy.
Moreover, Our final method (SMEM+JSD+Entropy) shows the best performance among the variants of our methods in all the Active Learning stages. Note that as Entropy here is the same as \emph{main entropy}, the Entropy curve is identical to that in \Fref{fig:quantitative}.}
\label{fig:ablation}
\end{figure}


\noindent\textbf{Comparison with existing approaches.}
In \Fref{fig:quantitative},
among the existing methods, \textbf{\emph{Margin}} and \textbf{\emph{Least Confident(LC)}} show similarly poor performance, while \textbf{\emph{Random}} and \textbf{\emph{Entropy}} show better performance in the early stage and the late stage respectively. 
We conjecture that the poor performance of \textbf{\emph{Margin}} and \textbf{\emph{LC}} is due to the large number of classes. VQA v2 has 3,129 candidate classes,
causing a long-tailed distribution problem, and makes conventional AL approaches difficult to select informative samples. In particular, the model predictions might be dominated to major classes; regarding only the most confident class (\textbf{\emph{LC}}) or second (\textbf{\emph{Margin}}) might lead to biased acquisition to major classes.
Note that recent state-of-the-art Active Learning methods, \textbf{\emph{LL}}, \textbf{\emph{VAAL}}, \textbf{\emph{VRAAL}}, show similar performance to \textbf{\emph{Random}} and \textbf{\emph{Entropy}} in the VQA task.
Moreover, our proposed method shows the best performance compared to all the counterparts by a large margin in all stages.
Note that the best among the existing approaches achieved 61.59\% accuracy when using 400K training samples, 
which is similar performance of our model trained with 280K training samples, signifying that SMEM can save about 30\% of the human labeling effort compared to existing approaches.

\noindent\textbf{Ablation study.}
We also perform an ablation study on each of the  components of our final method.
\textbf{\emph{+Distillation}} is for the proposed self-distillation loss 
in \Eref{eqn:loss_q}.
\textbf{\emph{+Jensen-Shannon Divergence (+JSD)}} is the additional JSD term which changes our acquisition function from \Eref{eqn:smem} to \Eref{eqn:addJSD}.
\textbf{\emph{+Entropy}} indicates the additional entropy term of the ``main'' branch output $H(\hat{Y})$.
In \Fref{fig:ablation} (b), we include traditional AL approaches, \textbf{\emph{Random}} and \textbf{\emph{Entropy}} for comparison. 
In \Fref{fig:ablation} (a), compared to the baselines with only single-modal entropy ($\alpha=0$ for V-only and $\alpha=1$ for Q-only in \Eref{eqn:smem}), using both V and Q modalities (\Eref{eqn:smem}) along with \textbf{\emph{+Distillation}} shows the best performance in all the AL stages.
In \Fref{fig:ablation} (b), SMEM without any auxiliary terms already shows favorable performance against traditional Active Learning methods, \textbf{\emph{Random}} and \textbf{\emph{Entropy}}.
Moreover, \textbf{\emph{+JSD}} noticeably improves the performance of SMEM, and our final method with both \textbf{\emph{+JSD}} and \textbf{\emph{+Entropy}} shows the best performance among all the variants of our methods in all AL stages.

\begin{figure}[t]
\centering
\begin{tabular}[c]{c}
    \subfigure[]{\includegraphics[width=.5\linewidth,keepaspectratio]{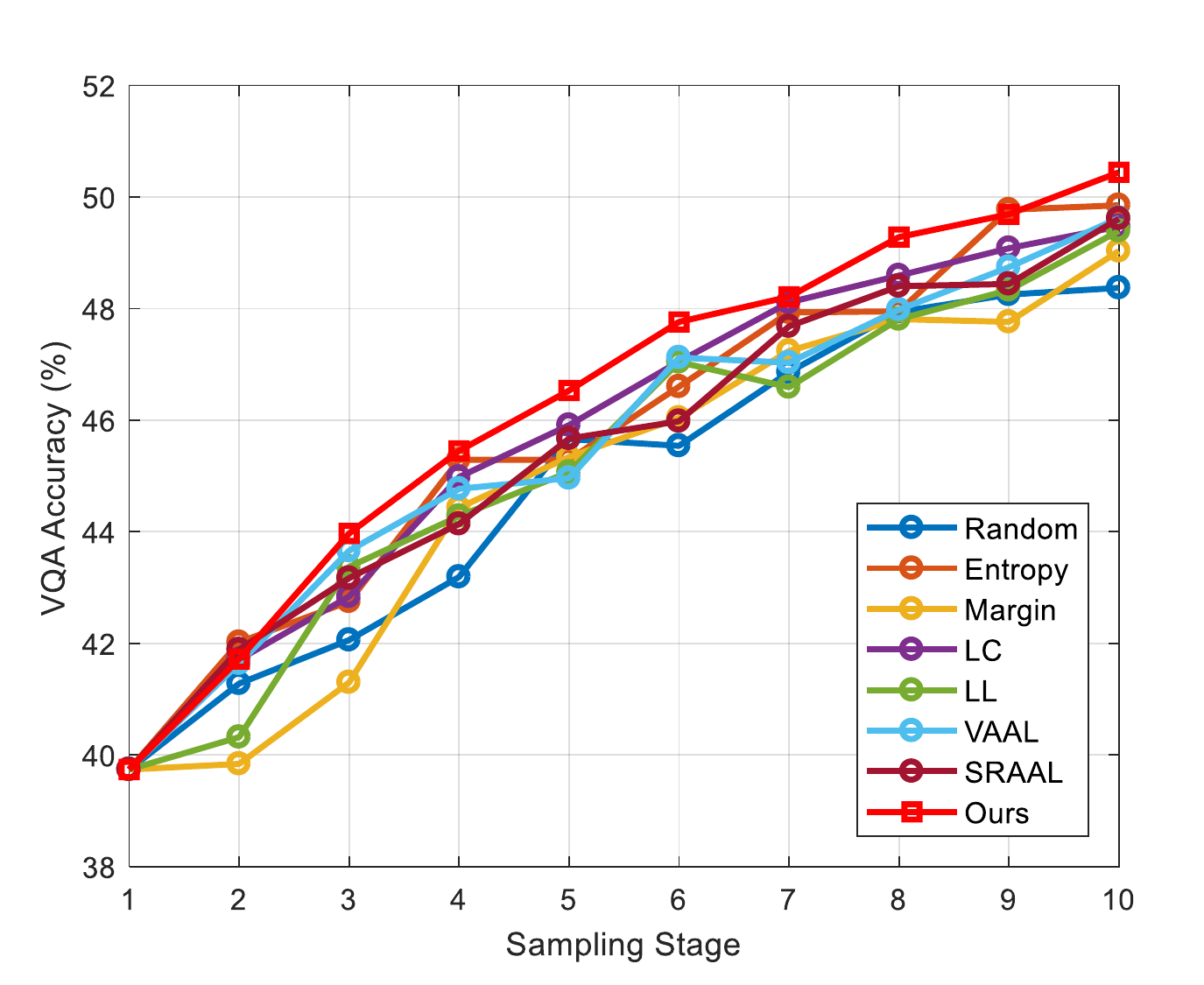}}			
    \subfigure[]{\includegraphics[width=.5\linewidth,keepaspectratio]{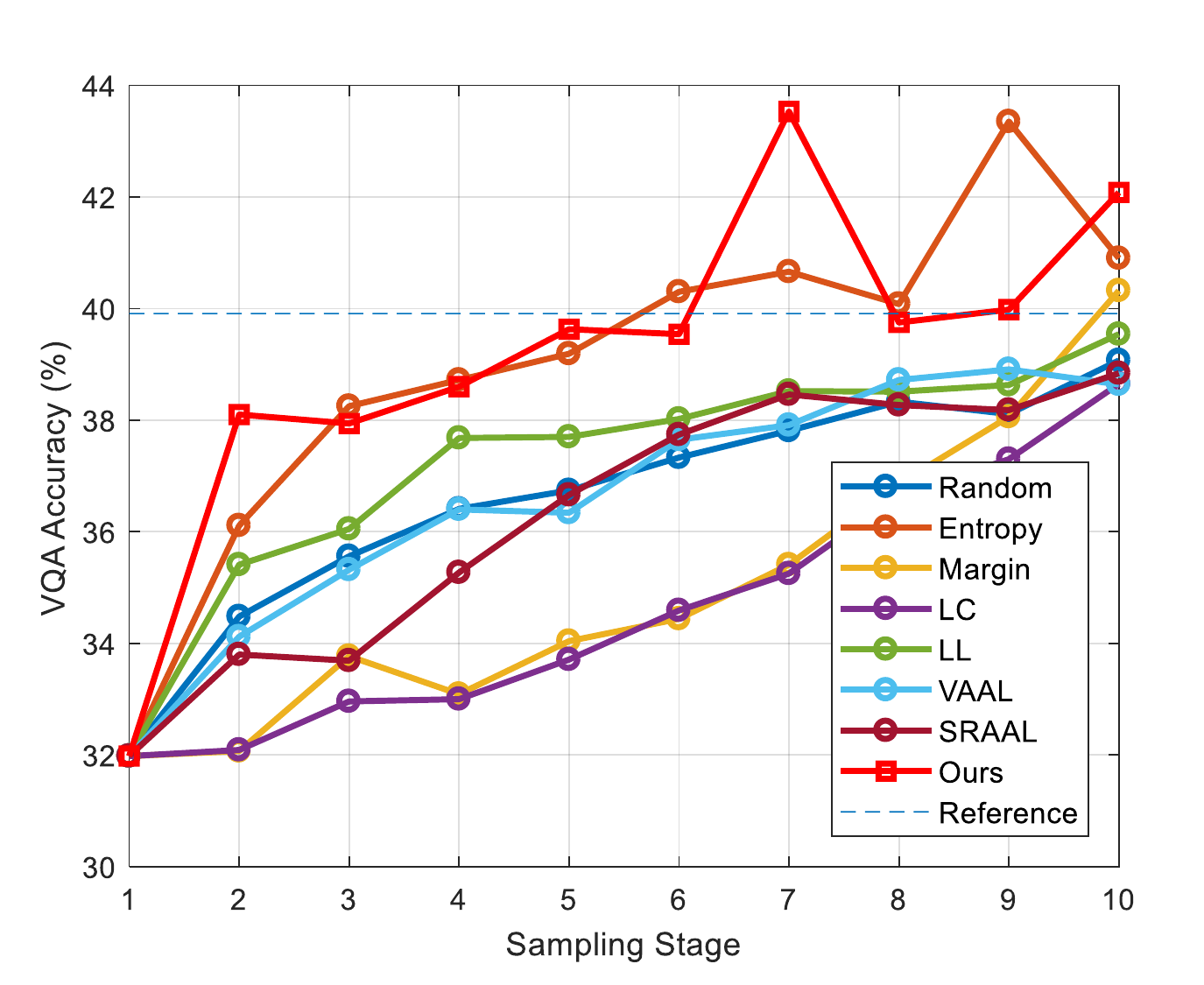}}
\end{tabular}
\caption{Comparison with existing approaches on the (a) VizWiz and (b) VQA-CP2 datasets. Our proposed method shows favorable performance on both datasets. The Reference line for (b) denotes the performance of the VQA model with the full 100\% of the dataset at 39.91\% accuracy.
} 
\label{fig:viz_cp}
\end{figure}


\begin{figure}[ht]
\vspace{0mm}
\centering
   \includegraphics[width=.5\linewidth]{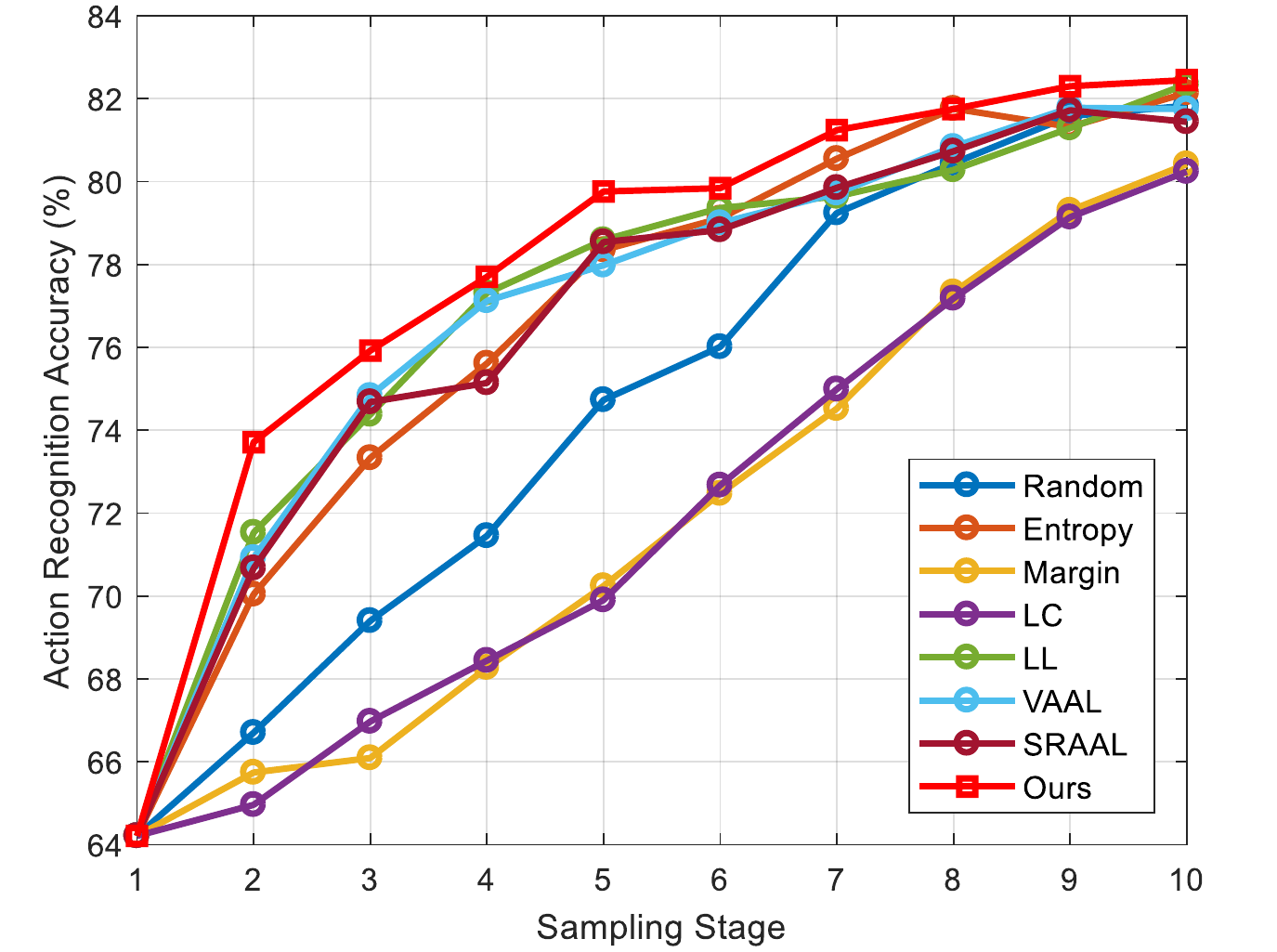}
   \caption{Comparison with existing approaches on the NTU RGB+D Action Recognition Dataset. Although our method is designed specifically for VQA task, our method consistently outperforms all the existing approaches in every sampling stage.
   }
\label{fig:NTU}
\end{figure}

\subsection{Experiments on Additional VQA Datasets: VizWiz and VQA-CP2}

To demonstrate the extensibility of our method, we extend our method on other VQA datasets.

\noindent\textbf{VizWiz VQA}~\cite{gurari2018vizwiz}, an arguably more difficult dataset, contains 19,425 train image/question pairs. We sample 1,900 image/question pairs per stage with the final stage ending at 19,000 image/question pairs.
For architecture, we use a publicly available code of~\cite{yang2016stacked}
, and show that our method is model agnostic.
We compare to the same baselines as in the previous subsection, and experimental results are shown in \Fref{fig:viz_cp} (a).
As VizWiz is a more challenging dataset than VQA v2 dataset, 
all the baselines show similar performance. Even the recent state-of-the-art such as \textbf{\emph{VAAL}} and \textbf{\emph{SRAAL}} show performance similar to that of all other baselines with no baseline being clearly above another.
Among all other baselines, our proposed method shows the best performance in most of the sampling stages.


\noindent\textbf{VQA-CP2}~\cite{goyal2017making} is a re-ordering of VQA v2
and the dataset size almost the same, containing 438K train image/question pairs. We sample 40K image/question pairs per stage with the final stage ending at 400K image/question pairs.
We use the same network architecture as VQA v2 and compare to the same baselines.
The experimental results are shown in \Fref{fig:viz_cp}(b).
Among the existing approaches, \textbf{\emph{Entropy}} shows the best performance, and the recent state-of-the-art methods of \textbf{\emph{VAAL}} and \textbf{\emph{SRAAL}} show mediocre performance, performing worse than \textbf{\emph{Random}} in several stages.
Note that when tested on all 438K training data, the model performance is 39.91\% (denoted as Reference in \Fref{fig:viz_cp}(b)) which is lower than the performance of \emph{\textbf{Entropy}} and \emph{\textbf{Ours} trained with the subset of the training data}. 
We find that by elaborately sampling some meaningful data, we are able to surpass the performance of the full dataset. 
As the semi-supervised learning for bias data started to be explored~\cite{kim2020distribution,oh2021distribution}, we believe this finding also opens up interesting possible future research ideas.

\subsection{Experiments on NTU Action Recognition Dataset}
Although we designed our model specifically for multi-modal task of VQA,
we also evaluate our model on one of the popular multi-modal action recognition datasets, NTU RGB+D dataset~\cite{shahroudy2016ntu}, to show the extensibility of our approach.
We use the model from~\cite{li2018co} with the NTU RGB+D dataset~\cite{shahroudy2016ntu} which leverages multiple views or subjects with the same action class and trains the network.
The dataset processing used in the implementation of~\cite{li2018co} makes use of 37,646 pairs of inputs and we sample 3,000 at each stage, ending at 30,000 pairs at the last stage. 
We use cross-subject evaluation~\cite{shahroudy2016ntu} that splits the 40 subjects into training and testing sets.
We apply our method on the subject dataset and treat each subject as a single modality and train additional classifiers similar to the VQA settings. 

We use the same baselines as the previous subsection, and show experimental results in \Fref{fig:NTU}.
Note that the gap between \textbf{\emph{Margin}} or \textbf{\emph{LC}} and the other baselines is larger than that of the previous experiments.
Among \textbf{\emph{Random}}, \textbf{\emph{Entropy}}, \textbf{\emph{LL}}, \textbf{\emph{VAAL}}, and \textbf{\emph{SRAAL}},
\textbf{\emph{LL}} and \textbf{\emph{VAAL}} shows the best performance in the early stages, and \textbf{\emph{Entropy}} shows the best performance in the later stages.
Moreover, compared to baseline approaches, SMEM shows favorable performance 
for all sampling stages in the NTU RGB+D dataset as well.
The effectiveness of SMEM opens up the possibilities of applying our method on other datasets and tasks where multiple inputs maybe used whether they are of the same modality of not.

\section{Conclusion}
In this work, we introduced a novel Active Learning framework specifically tailored for the multi-modal task of Visual Question Answering. We showed through our empirical evidence the performance gains of our novel method 
in the Active Learning setup compared to prior state-of-the-art Active Learning methods. We also empirically showed that our method is extendable to other multi-modal applications such as Action Recognition with ease and promise, which opens a window for further research in other 
tasks. 
In addition, we also show through our testing on the VQA-CP2 dataset that by using our method to sample a subset of the data
surpasses using the full dataset, showing that a higher performance can be achieved with less data. 
We believe this 
shows the potential possibilities of our research and shed new light on this field and other multi-modal research fields. We hope to explore Active Learning for other multi-modal tasks~\cite{cho2021dealing,johnson2016densecap,kim2019dense,kim2021dense} and encourage future researchers to study Active Learning for other multi-modal tasks in the future as well.

\clearpage

\noindent{\textbf{Acknowledgements.}}
This work was supported by the Institute for Information \& Communications Technology Promotion (2017-0-01772) grant funded by the Korea government.

\bibliography{egbib}

\begin{thebibliography}{61}
\providecommand{\natexlab}[1]{#1}
\providecommand{\url}[1]{\texttt{#1}}
\expandafter\ifx\csname urlstyle\endcsname\relax
  \providecommand{\doi}[1]{doi: #1}\else
  \providecommand{\doi}{doi: \begingroup \urlstyle{rm}\Url}\fi

\bibitem[Anderson et~al.(2018)Anderson, He, Buehler, Teney, Johnson, Gould, and
  Zhang]{anderson2018bottom}
Peter Anderson, Xiaodong He, Chris Buehler, Damien Teney, Mark Johnson, Stephen
  Gould, and Lei Zhang.
\newblock Bottom-up and top-down attention for image captioning and visual
  question answering.
\newblock In \emph{{IEEE} Conference on Computer Vision and Pattern Recognition
  (CVPR)}, 2018.

\bibitem[Antol et~al.(2015)Antol, Agrawal, Lu, Mitchell, Batra,
  Lawrence~Zitnick, and Parikh]{antol2015vqa}
Stanislaw Antol, Aishwarya Agrawal, Jiasen Lu, Margaret Mitchell, Dhruv Batra,
  C~Lawrence~Zitnick, and Devi Parikh.
\newblock Vqa: Visual question answering.
\newblock In \emph{{IEEE} International Conference on Computer Vision (ICCV)},
  2015.

\bibitem[Cadene et~al.(2019)Cadene, Dancette, Ben-younes, Cord, and
  Parikh]{cadene2019rubi}
Remi Cadene, Corentin Dancette, Hedi Ben-younes, Matthieu Cord, and Devi
  Parikh.
\newblock Rubi: Reducing unimodal biases in visual question answering.
\newblock In \emph{Advances in Neural Information Processing Systems (NIPS)},
  2019.

\bibitem[Chi et~al.(2019)Chi, Eric, Kim, Shen, and
  Hakkani-tur]{chi2019justasknavigation}
Ta-Chung Chi, Mihail Eric, Seokhwan Kim, Minmin Shen, and Dilek Hakkani-tur.
\newblock Just ask:an interactive learning framework for vision and language
  navigation, 2019.

\bibitem[Cho et~al.(2021{\natexlab{a}})Cho, Kim, Choi, Jung, and
  Kweon]{cho2021dealing}
Jae~Won Cho, Dong-Jin Kim, Jinsoo Choi, Yunjae Jung, and In~So Kweon.
\newblock Dealing with missing modalities in the visual question
  answer-difference prediction task through knowledge distillation.
\newblock In \emph{{IEEE} Conference on Computer Vision and Pattern Recognition
  (CVPR)}, 2021{\natexlab{a}}.

\bibitem[Cho et~al.(2021{\natexlab{b}})Cho, Kim, Jung, and Kweon]{cho2021mcdal}
Jae~Won Cho, Dong-Jin Kim, Yunjae Jung, and In~So Kweon.
\newblock Mcdal: Maximum classifier discrepancy for active learning.
\newblock \emph{arXiv preprint arXiv:2107.11049}, 2021{\natexlab{b}}.

\bibitem[Cohn et~al.(1996)Cohn, Ghahramani, and Jordan]{cohn1996active}
David~A Cohn, Zoubin Ghahramani, and Michael~I Jordan.
\newblock Active learning with statistical models.
\newblock \emph{Journal of artificial intelligence research}, 4:\penalty0
  129--145, 1996.

\bibitem[Collins et~al.(2008)Collins, Deng, Li, and
  Fei-Fei]{collins2008towards}
Brendan Collins, Jia Deng, Kai Li, and Li~Fei-Fei.
\newblock Towards scalable dataset construction: An active learning approach.
\newblock In \emph{European Conference on Computer Vision (ECCV)}, 2008.

\bibitem[Culotta and McCallum(2005)]{culotta2005reducing}
Aron Culotta and Andrew McCallum.
\newblock Reducing labeling effort for structured prediction tasks.
\newblock In \emph{AAAI Conference on Artificial Intelligence (AAAI)}, 2005.

\bibitem[Deng et~al.(2009)Deng, Dong, Socher, Li, Li, and
  Fei-Fei]{deng2009imagenet}
Jia Deng, Wei Dong, Richard Socher, Li-Jia Li, Kai Li, and Li~Fei-Fei.
\newblock Imagenet: A large-scale hierarchical image database.
\newblock In \emph{{IEEE} Conference on Computer Vision and Pattern Recognition
  (CVPR)}, 2009.

\bibitem[Figueroa et~al.(2012)Figueroa, Zeng-Treitler, Ngo, Goryachev, and
  Wiechmann]{figueroa2012active}
Rosa~L Figueroa, Qing Zeng-Treitler, Long~H Ngo, Sergey Goryachev, and
  Eduardo~P Wiechmann.
\newblock Active learning for clinical text classification: is it better than
  random sampling?
\newblock \emph{Journal of the American Medical Informatics Association},
  19\penalty0 (5):\penalty0 809--816, 2012.

\bibitem[Fukui et~al.(2016)Fukui, Park, Yang, Rohrbach, Darrell, and
  Rohrbach]{fukui2016multimodal}
Akira Fukui, Dong~Huk Park, Daylen Yang, Anna Rohrbach, Trevor Darrell, and
  Marcus Rohrbach.
\newblock Multimodal compact bilinear pooling for visual question answering and
  visual grounding.
\newblock In \emph{Conference on Empirical Methods in Natural Language
  Processing (EMNLP)}, 2016.

\bibitem[Gal et~al.(2017)Gal, Islam, and Ghahramani]{gal2017deep}
Yarin Gal, Riashat Islam, and Zoubin Ghahramani.
\newblock Deep bayesian active learning with image data.
\newblock In \emph{International Conference on Machine Learning (ICML)}, 2017.

\bibitem[Goyal et~al.(2017)Goyal, Khot, Summers-Stay, Batra, and
  Parikh]{goyal2017making}
Yash Goyal, Tejas Khot, Douglas Summers-Stay, Dhruv Batra, and Devi Parikh.
\newblock Making the v in vqa matter: Elevating the role of image understanding
  in visual question answering.
\newblock In \emph{{IEEE} Conference on Computer Vision and Pattern Recognition
  (CVPR)}, 2017.

\bibitem[Gurari et~al.(2018)Gurari, Li, Stangl, Guo, Lin, Grauman, Luo, and
  Bigham]{gurari2018vizwiz}
Danna Gurari, Qing Li, Abigale~J Stangl, Anhong Guo, Chi Lin, Kristen Grauman,
  Jiebo Luo, and Jeffrey~P Bigham.
\newblock Vizwiz grand challenge: Answering visual questions from blind people.
\newblock In \emph{{IEEE} Conference on Computer Vision and Pattern Recognition
  (CVPR)}, 2018.

\bibitem[Hinton et~al.(2015)Hinton, Vinyals, and Dean]{hinton2015distilling}
Geoffrey Hinton, Oriol Vinyals, and Jeff Dean.
\newblock Distilling the knowledge in a neural network.
\newblock \emph{arXiv preprint arXiv:1503.02531}, 2015.

\bibitem[Hoffman et~al.(2016)Hoffman, Gupta, and Darrell]{hoffman2016learning}
Judy Hoffman, Saurabh Gupta, and Trevor Darrell.
\newblock Learning with side information through modality hallucination.
\newblock In \emph{{IEEE} Conference on Computer Vision and Pattern Recognition
  (CVPR)}, 2016.

\bibitem[Hoi et~al.(2006)Hoi, Jin, Zhu, and Lyu]{hoi2006batch}
Steven~CH Hoi, Rong Jin, Jianke Zhu, and Michael~R Lyu.
\newblock Batch mode active learning and its application to medical image
  classification.
\newblock In \emph{International Conference on Machine Learning (ICML)}, 2006.

\bibitem[Houlsby et~al.(2011)Houlsby, Husz{\'a}r, Ghahramani, and
  Lengyel]{houlsby2011bayesian}
Neil Houlsby, Ferenc Husz{\'a}r, Zoubin Ghahramani, and M{\'a}t{\'e} Lengyel.
\newblock Bayesian active learning for classification and preference learning.
\newblock \emph{arXiv preprint arXiv:1112.5745}, 2011.

\bibitem[Johnson et~al.(2016)Johnson, Karpathy, and
  Fei-Fei]{johnson2016densecap}
Justin Johnson, Andrej Karpathy, and Li~Fei-Fei.
\newblock Densecap: Fully convolutional localization networks for dense
  captioning.
\newblock In \emph{{IEEE} Conference on Computer Vision and Pattern Recognition
  (CVPR)}, 2016.

\bibitem[Johnson et~al.(2017)Johnson, Hariharan, van~der Maaten, Fei-Fei,
  Lawrence~Zitnick, and Girshick]{johnson2017clevr}
Justin Johnson, Bharath Hariharan, Laurens van~der Maaten, Li~Fei-Fei,
  C~Lawrence~Zitnick, and Ross Girshick.
\newblock Clevr: A diagnostic dataset for compositional language and elementary
  visual reasoning.
\newblock In \emph{{IEEE} Conference on Computer Vision and Pattern Recognition
  (CVPR)}, 2017.

\bibitem[Jones et~al.(2003)Jones, Ghani, Mitchell, and Riloff]{jones2003active}
Rosie Jones, Rayid Ghani, Tom Mitchell, and Ellen Riloff.
\newblock Active learning for information extraction with multiple view feature
  sets.
\newblock \emph{Proc. of Adaptive Text Extraction and Mining, EMCL/PKDD-03,
  Cavtat-Dubrovnik, Croatia}, 2003.

\bibitem[Joshi et~al.(2009)Joshi, Porikli, and
  Papanikolopoulos]{joshi2009multi}
Ajay~J Joshi, Fatih Porikli, and Nikolaos Papanikolopoulos.
\newblock Multi-class active learning for image classification.
\newblock In \emph{{IEEE} Conference on Computer Vision and Pattern Recognition
  (CVPR)}, 2009.

\bibitem[Kendall and Gal(2017)]{kendall2017uncertainties}
Alex Kendall and Yarin Gal.
\newblock What uncertainties do we need in bayesian deep learning for computer
  vision?
\newblock In \emph{Advances in Neural Information Processing Systems (NIPS)},
  2017.

\bibitem[Kim et~al.(2018{\natexlab{a}})Kim, Choi, Oh, Yoon, and
  Kweon]{kim2018disjoint}
Dong-Jin Kim, Jinsoo Choi, Tae-Hyun Oh, Youngjin Yoon, and In~So Kweon.
\newblock Disjoint multi-task learning between heterogeneous human-centric
  tasks.
\newblock In \emph{IEEE Winter Conference on Applications of Computer Vision
  (WACV)}, 2018{\natexlab{a}}.

\bibitem[Kim et~al.(2019{\natexlab{a}})Kim, Choi, Oh, and Kweon]{kim2019dense}
Dong-Jin Kim, Jinsoo Choi, Tae-Hyun Oh, and In~So Kweon.
\newblock Dense relational captioning: Triple-stream networks for
  relationship-based captioning.
\newblock In \emph{{IEEE} Conference on Computer Vision and Pattern Recognition
  (CVPR)}, 2019{\natexlab{a}}.

\bibitem[Kim et~al.(2019{\natexlab{b}})Kim, Choi, Oh, and Kweon]{kim2019image}
Dong-Jin Kim, Jinsoo Choi, Tae-Hyun Oh, and In~So Kweon.
\newblock Image captioning with very scarce supervised data: Adversarial
  semi-supervised learning approach.
\newblock In \emph{Conference on Empirical Methods in Natural Language
  Processing (EMNLP)}, 2019{\natexlab{b}}.

\bibitem[Kim et~al.(2020{\natexlab{a}})Kim, Sun, Choi, Lin, and
  Kweon]{kim2020detecting}
Dong-Jin Kim, Xiao Sun, Jinsoo Choi, Stephen Lin, and In~So Kweon.
\newblock Detecting human-object interactions with action co-occurrence priors.
\newblock In \emph{European Conference on Computer Vision (ECCV)},
  2020{\natexlab{a}}.

\bibitem[Kim et~al.(2021{\natexlab{a}})Kim, Oh, Choi, and Kweon]{kim2021dense}
Dong-Jin Kim, Tae-Hyun Oh, Jinsoo Choi, and In~So Kweon.
\newblock Dense relational image captioning via multi-task triple-stream
  networks.
\newblock \emph{{IEEE} Transactions on Pattern Analysis and Machine
  Intelligence (TPAMI)}, 2021{\natexlab{a}}.

\bibitem[Kim et~al.(2021{\natexlab{b}})Kim, Sun, Choi, Lin, and
  Kweon]{kim2021acp++}
Dong-Jin Kim, Xiao Sun, Jinsoo Choi, Stephen Lin, and In~So Kweon.
\newblock Acp++: Action co-occurrence priors for human-object interaction
  detection.
\newblock \emph{{IEEE} Transactions on Image Processing (TIP)},
  2021{\natexlab{b}}.

\bibitem[Kim et~al.(2020{\natexlab{b}})Kim, Hur, Park, Yang, Hwang, and
  Shin]{kim2020distribution}
Jaehyung Kim, Youngbum Hur, Sejun Park, Eunho Yang, Sung~Ju Hwang, and Jinwoo
  Shin.
\newblock Distribution aligning refinery of pseudo-label for imbalanced
  semi-supervised learning.
\newblock In \emph{Advances in Neural Information Processing Systems (NIPS)},
  2020{\natexlab{b}}.

\bibitem[Kim et~al.(2018{\natexlab{b}})Kim, Jun, and Zhang]{kim2018bilinear}
Jin-Hwa Kim, Jaehyun Jun, and Byoung-Tak Zhang.
\newblock Bilinear attention networks.
\newblock In \emph{Advances in Neural Information Processing Systems (NIPS)},
  2018{\natexlab{b}}.

\bibitem[Kingma and Welling(2013)]{kingma2013auto}
Diederik~P Kingma and Max Welling.
\newblock Auto-encoding variational bayes.
\newblock \emph{arXiv preprint arXiv:1312.6114}, 2013.

\bibitem[Krishna et~al.(2017)Krishna, Zhu, Groth, Johnson, Hata, Kravitz, Chen,
  Kalantidis, Li, Shamma, et~al.]{krishna2017visual}
Ranjay Krishna, Yuke Zhu, Oliver Groth, Justin Johnson, Kenji Hata, Joshua
  Kravitz, Stephanie Chen, Yannis Kalantidis, Li-Jia Li, David~A Shamma, et~al.
\newblock Visual genome: Connecting language and vision using crowdsourced
  dense image annotations.
\newblock \emph{International Journal of Computer Vision (IJCV)}, 123\penalty0
  (1):\penalty0 32--73, 2017.

\bibitem[Lewis and Gale(1994)]{lewis1994sequential}
David~D Lewis and William~A Gale.
\newblock A sequential algorithm for training text classifiers.
\newblock In \emph{SIGIR’94}, 1994.

\bibitem[Li et~al.(2018)Li, Zhong, Xie, and Pu]{li2018co}
Chao Li, Qiaoyong Zhong, Di~Xie, and Shiliang Pu.
\newblock Co-occurrence feature learning from skeleton data for action
  recognition and detection with hierarchical aggregation.
\newblock In \emph{International Joint Conference on Artificial Intelligence
  (IJCAI)}, 2018.

\bibitem[Li and Hoiem(2016)]{li2016learning}
Zhizhong Li and Derek Hoiem.
\newblock Learning without forgetting.
\newblock In \emph{European Conference on Computer Vision (ECCV)}, 2016.

\bibitem[Lin and Parikh(2017)]{lin2017activelearningvqa}
Xiao Lin and Devi Parikh.
\newblock Active learning for visual question answering: An empirical study,
  2017.

\bibitem[Liu et~al.(2018)Liu, Li, Shao, Chen, and Wang]{liu2018show}
Xihui Liu, Hongsheng Li, Jing Shao, Dapeng Chen, and Xiaogang Wang.
\newblock Show, tell and discriminate: Image captioning by self-retrieval with
  partially labeled data.
\newblock In \emph{European Conference on Computer Vision (ECCV)}, 2018.

\bibitem[Lu et~al.(2016)Lu, Yang, Batra, and Parikh]{lu2016hierarchical}
Jiasen Lu, Jianwei Yang, Dhruv Batra, and Devi Parikh.
\newblock Hierarchical question-image co-attention for visual question
  answering.
\newblock In \emph{Advances in Neural Information Processing Systems (NIPS)},
  2016.

\bibitem[Mairesse et~al.(2010)Mairesse, Ga{\v{s}}i{\'c},
  Jur{\v{c}}{\'\i}{\v{c}}ek, Keizer, Thomson, Yu, and
  Young]{mairesse2010phrase}
Fran{\c{c}}ois Mairesse, Milica Ga{\v{s}}i{\'c}, Filip
  Jur{\v{c}}{\'\i}{\v{c}}ek, Simon Keizer, Blaise Thomson, Kai Yu, and Steve
  Young.
\newblock Phrase-based statistical language generation using graphical models
  and active learning.
\newblock In \emph{Annual Meeting of the Association for Computational
  Linguistics (ACL)}, 2010.

\bibitem[Misra et~al.(2018)Misra, Girshick, Fergus, Hebert, Gupta, and Van
  Der~Maaten]{misra2018learning}
Ishan Misra, Ross Girshick, Rob Fergus, Martial Hebert, Abhinav Gupta, and
  Laurens Van Der~Maaten.
\newblock Learning by asking questions.
\newblock In \emph{{IEEE} Conference on Computer Vision and Pattern Recognition
  (CVPR)}, 2018.

\bibitem[Oh et~al.(2021)Oh, Kim, and Kweon]{oh2021distribution}
Youngtaek Oh, Dong-Jin Kim, and In~So Kweon.
\newblock Distribution-aware semantics-oriented pseudo-label for imbalanced
  semi-supervised learning.
\newblock \emph{arXiv preprint arXiv:2106.05682}, 2021.

\bibitem[Phuong and Lampert(2019)]{phuong2019distillation}
Mary Phuong and Christoph~H Lampert.
\newblock Distillation-based training for multi-exit architectures.
\newblock In \emph{{IEEE} International Conference on Computer Vision (ICCV)},
  2019.

\bibitem[Scheffer et~al.(2001)Scheffer, Decomain, and
  Wrobel]{scheffer2001active}
Tobias Scheffer, Christian Decomain, and Stefan Wrobel.
\newblock Active hidden markov models for information extraction.
\newblock In \emph{International Symposium on Intelligent Data Analysis}, 2001.

\bibitem[Sener and Savarese(2017)]{sener2017active}
Ozan Sener and Silvio Savarese.
\newblock Active learning for convolutional neural networks: A core-set
  approach.
\newblock \emph{arXiv preprint arXiv:1708.00489}, 2017.

\bibitem[Settles(2009)]{settles2009active}
Burr Settles.
\newblock Active learning literature survey.
\newblock Technical report, University of Wisconsin-Madison Department of
  Computer Sciences, 2009.

\bibitem[Settles et~al.(2008)Settles, Craven, and Friedland]{settles2008active}
Burr Settles, Mark Craven, and Lewis Friedland.
\newblock Active learning with real annotation costs.
\newblock In \emph{Proceedings of the NIPS workshop on cost-sensitive
  learning}, 2008.

\bibitem[Shahroudy et~al.(2016)Shahroudy, Liu, Ng, and Wang]{shahroudy2016ntu}
Amir Shahroudy, Jun Liu, Tian-Tsong Ng, and Gang Wang.
\newblock Ntu rgb+ d: A large scale dataset for 3d human activity analysis.
\newblock In \emph{{IEEE} Conference on Computer Vision and Pattern Recognition
  (CVPR)}, 2016.

\bibitem[Shen et~al.(2019)Shen, Kar, and Fidler]{shen2019learning}
Tingke Shen, Amlan Kar, and Sanja Fidler.
\newblock Learning to caption images through a lifetime by asking questions.
\newblock In \emph{{IEEE} International Conference on Computer Vision (ICCV)},
  2019.

\bibitem[Shen et~al.(2017)Shen, Yun, Lipton, Kronrod, and
  Anandkumar]{shen2017deep}
Yanyao Shen, Hyokun Yun, Zachary~C Lipton, Yakov Kronrod, and Animashree
  Anandkumar.
\newblock Deep active learning for named entity recognition.
\newblock \emph{arXiv preprint arXiv:1707.05928}, 2017.

\bibitem[Shin et~al.(2021)Shin, Kim, Cho, Woo, Park, and Kweon]{shin2021labor}
Inkyu Shin, Dong-Jin Kim, Jae~Won Cho, Sanghyun Woo, KwanYong Park, and In~So
  Kweon.
\newblock Labor: Labeling only if required for domain adaptive semantic
  segmentation.
\newblock In \emph{{IEEE} International Conference on Computer Vision (ICCV)},
  2021.

\bibitem[Sinha et~al.(2019)Sinha, Ebrahimi, and Darrell]{sinha2019variational}
Samarth Sinha, Sayna Ebrahimi, and Trevor Darrell.
\newblock Variational adversarial active learning.
\newblock In \emph{{IEEE} International Conference on Computer Vision (ICCV)},
  2019.

\bibitem[Teney and van~den Hengel(2018)]{teney2018visual}
Damien Teney and Anton van~den Hengel.
\newblock Visual question answering as a meta learning task.
\newblock In \emph{European Conference on Computer Vision (ECCV)}, 2018.

\bibitem[Tong and Koller(2001)]{tong2001support}
Simon Tong and Daphne Koller.
\newblock Support vector machine active learning with applications to text
  classification.
\newblock \emph{Journal of Machine Learning Research (JMLR)}, 2\penalty0
  (Nov):\penalty0 45--66, 2001.

\bibitem[Wang et~al.(2020)Wang, Li, Ma, Ma, Guan, and Zheng]{shuo2020daal}
Shuo Wang, Yuexiang Li, Kai Ma, Ruhui Ma, Haibing Guan, and Yefeng Zheng.
\newblock Dual adversarial network for deep active learning.
\newblock In \emph{European Conference on Computer Vision (ECCV)}, 2020.

\bibitem[Yang et~al.(2016)Yang, He, Gao, Deng, and Smola]{yang2016stacked}
Zichao Yang, Xiaodong He, Jianfeng Gao, Li~Deng, and Alex Smola.
\newblock Stacked attention networks for image question answering.
\newblock In \emph{{IEEE} Conference on Computer Vision and Pattern Recognition
  (CVPR)}, 2016.

\bibitem[Yoo and Kweon(2019)]{yoo2019learning}
Donggeun Yoo and In~So Kweon.
\newblock Learning loss for active learning.
\newblock In \emph{{IEEE} Conference on Computer Vision and Pattern Recognition
  (CVPR)}, 2019.

\bibitem[Zhang et~al.(2020)Zhang, Li, Yang, Wang, Zha, and
  Huang]{zhang2020state}
Beichen Zhang, Liang Li, Shijie Yang, Shuhui Wang, Zheng-Jun Zha, and Qingming
  Huang.
\newblock State-relabeling adversarial active learning.
\newblock In \emph{{IEEE} Conference on Computer Vision and Pattern Recognition
  (CVPR)}, 2020.

\bibitem[Zhang et~al.(2019)Zhang, Song, Gao, Chen, Bao, and Ma]{zhang2019your}
Linfeng Zhang, Jiebo Song, Anni Gao, Jingwei Chen, Chenglong Bao, and Kaisheng
  Ma.
\newblock Be your own teacher: Improve the performance of convolutional neural
  networks via self distillation.
\newblock In \emph{{IEEE} International Conference on Computer Vision (ICCV)},
  2019.

\bibitem[Zhu et~al.(2016)Zhu, Groth, Bernstein, and Fei-Fei]{zhu2016visual7w}
Yuke Zhu, Oliver Groth, Michael Bernstein, and Li~Fei-Fei.
\newblock Visual7w: Grounded question answering in images.
\newblock In \emph{{IEEE} Conference on Computer Vision and Pattern Recognition
  (CVPR)}, 2016.

\end{thebibliography}


\end{document}


\maketitle


\section{Supplementary}

The contents of this supplementary material include 
implementation details of our model,
additional qualitative results,  
and additional analysis which were not included in the main paper due to space limitations.

\begin{algorithm}[t]
\begin{algorithmic}
\INPUT{ Labeled pool $\mathcal{D}\textsubscript{L}$, Initialized models $g_v(\cdot), g_q(\cdot),\text{ and }F(\cdot,\cdot)$ with parameters $\theta_v$,$\theta_q$,\text{ and }$\theta$ respectively}\\
\INPUT{Hyper-parameters : learning rate $\eta$, maximum number of epoches $MaxEpoch$}\\
\OUTPUT{ $g_v(\cdot), g_q(\cdot),\text{ and }F(\cdot,\cdot)$ with trained $\theta_v$,$\theta_q$,\text{ and }$\theta$}
{
    \FOR{$e=1$ to MaxEpoch}{
    \STATE sample $(X,Y)\in\mathcal{D}\textsubscript{L}$\;
    \STATE Compute $\mathcal{L}_{main}$\;
    \STATE Compute $\mathcal{L}_v$\;
    \STATE Compute $\mathcal{L}_q$\;
    \STATE Update the model parameters:\\
    \STATE $\theta \leftarrow \theta - \eta \nabla \mathcal{L}_{main}$\\
    \STATE $\theta_v \leftarrow \theta_v - \eta \nabla \mathcal{L}_{v}$\\
    \STATE $\theta_q \leftarrow \theta_q - \eta \nabla \mathcal{L}_{q}$\\
    }
    \ENDFOR
}
\end{algorithmic}
\caption{Training Our Model with Self-distillation}
\label{alg:training}
\end{algorithm}

\begin{algorithm}[t]
\begin{algorithmic}
\INPUT{Labeled pool $\mathcal{D}\textsubscript{L}$, Unlabeled pool $\mathcal{D}\textsubscript{U}$, Number of samples collecting for each stage $b$, Models $g_v(\cdot), g_q(\cdot),\text{ and }F(\cdot,\cdot)$ }
\OUTPUT{Updated $\mathcal{D}\textsubscript{L}$ and $\mathcal{D}\textsubscript{U}$}
{
    \STATE From $\mathcal{D}\textsubscript{U}$, collect a set of samples $\{X^s\}$ with 
    \STATE $\max_b S(V,Q) = \max_b{ \alpha H(Y_q)} \;$
    \hfill\STATE $+ (1-\alpha) H(Y_v) + \beta JSD(\mathbf{y}_v||\mathbf{y}_q) + \gamma H(\hat{Y})$\;
    \STATE $\{(X^s,Y^s)\} = ORACLE(\{X^s\})$\;
    \STATE $\mathcal{D}\textsubscript{L} \leftarrow \mathcal{D}\textsubscript{L} \cup \{(X^s,Y^s)\}$\;
    \STATE $\mathcal{D}\textsubscript{U} \leftarrow \mathcal{D}\textsubscript{U} - \{X^s\}$\;
}
\end{algorithmic}
 \caption{Our Proposed Sampling Strategy with Single-Modal Entropic Measure (SMEM)}
 \label{alg:sampling}
\end{algorithm}

\begin{figure*}[t]
\centering
\subfigure[Vanilla]{\includegraphics[width=0.32\linewidth]{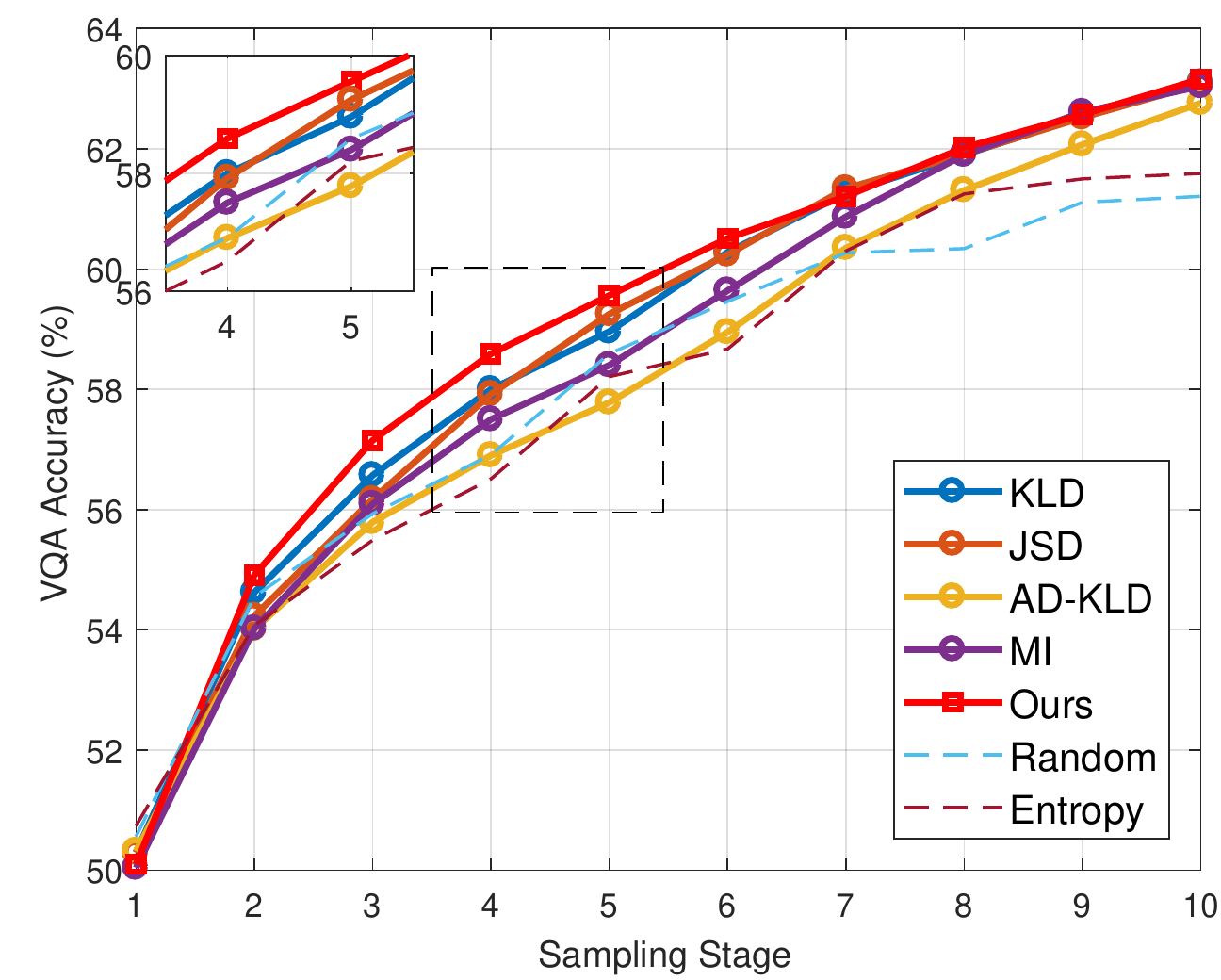}}
\subfigure[+Distillation]{\includegraphics[width=0.32\linewidth]{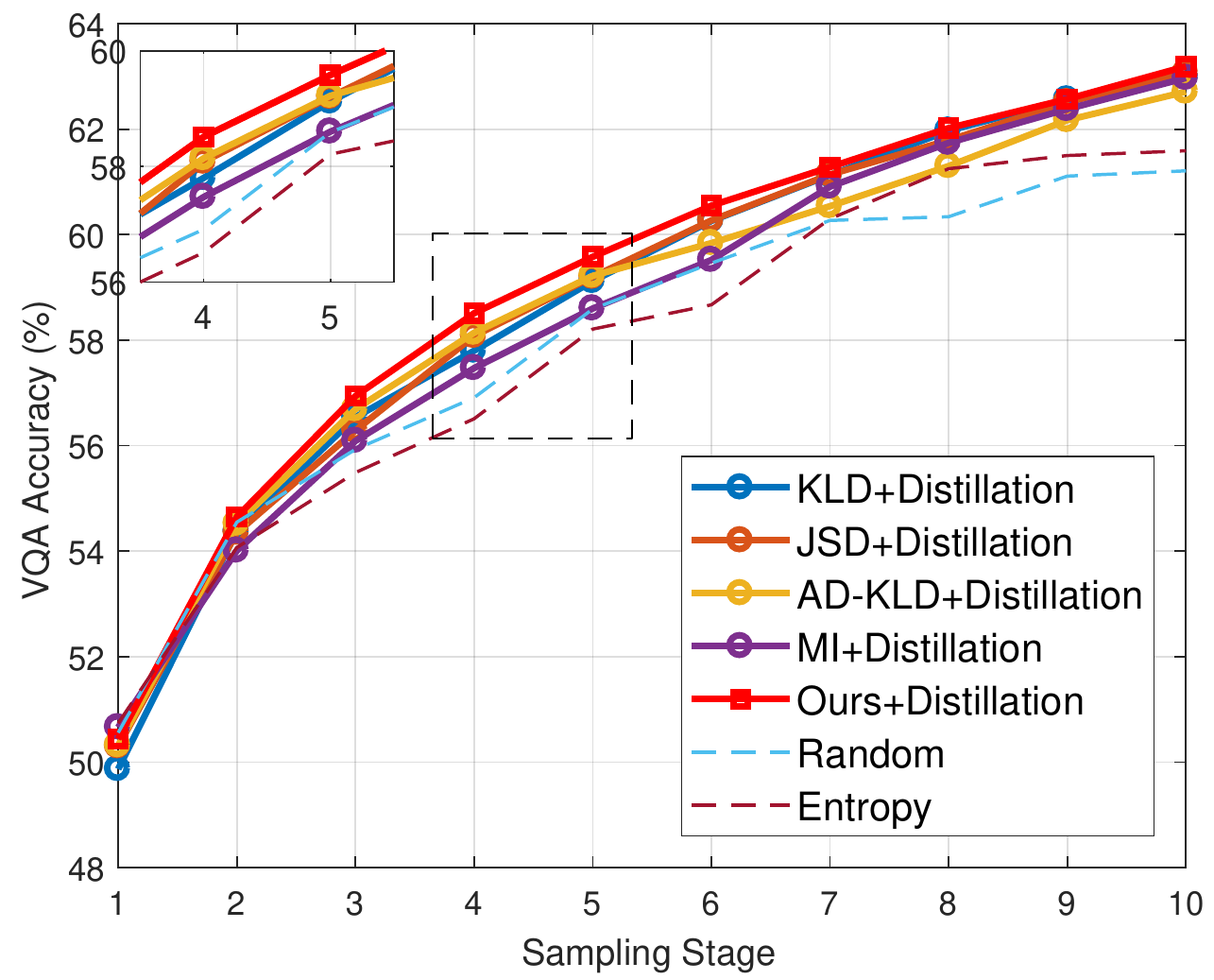}}
\subfigure[+Distillation+Entropy]{\includegraphics[width=0.32\linewidth]{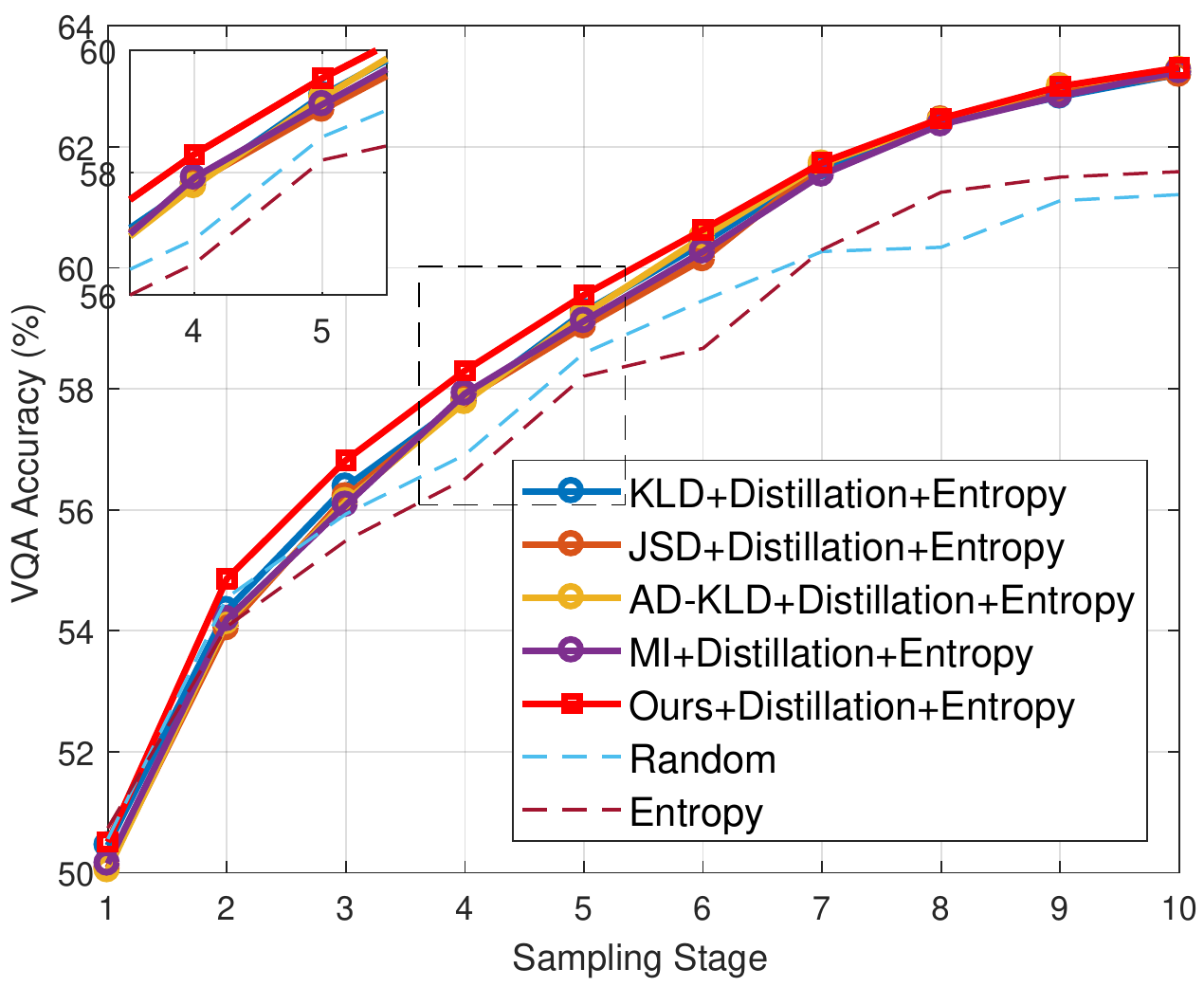}}
\caption{Comparison between different measuring methods for our method. 
Ours is the SMEM + JSD in all graphs above. 
(a):without anything, (b):with distillation loss, and (c):with both distillation loss and entropy.}
\label{fig:ablation}
\end{figure*}

\subsection{Implementation Details}
The VQA 2.0 Dataset can be found at this link (\url{https://visualqa.org/}).
Following the model from~\cite{anderson2018bottom}, we make use of the pre-extracted image features from Mask-RCNN~\cite{he2017mask} for the visual input and Glove6B~\cite{pennington2014glove} embedding with a GRU~\cite{chung2014empirical} to process the question features. Then, the features are attended using the modified version of the attention proposed in~\cite{anderson2018bottom}~\footnote{The modified version of the attention proposed in~\cite{anderson2018bottom} is found on this github link: https://github.com/hengyuan-hu/bottom-up-attention-vqa}.
We train our model with a batch size of 512 and use the Adamax optimizer from the PyTorch library. The hyperparameters used are: Learning Rate: 0.002, Betas: (0.9, 0.999), eps: 1$e^{-0.8}$, Weight Decay: 0.
Model hyperparameters are listed next. The vocabulary size used is 19902, the output of the Mask-RCNN is 2048, as for the Attention, the input is set at 2048 for the visual input, but all hidden states, question input, and output values are set to 1024. The word embedding size used is 300, and the GRU hidden state is set to 1024. The classifier, which can be seen as all 3 classifiers as stated in the method section of our paper, has an input of 1024 and an output of 3129, which is the number of answer candidates.

The VQA accuracy measure is from~\cite{antol2015vqa} and can be summarised as follows:
\begin{equation}
    Accuracy(A) = min\left\{\frac{\text{\# of humans that said }A}{3}, 1\right\},
    \label{eqn:VQA_accuracy}
\end{equation}
where $A$ is the Answer prediction from the model.


For our Active Learning setup, we set 10 stages with 40,000 question image pairs for each step, and increase 40,000 pairs each time, ending up at 400,000 at the final stage. We do not reinitialize the model each time but continue using the model that was previously trained. At Stage 0, we train the model for 20 epochs. For the next Stages from 1 to 9, as the model is already pre-trained in the previous stages, we train for only 10 epochs each. At the end of each epoch, the model is evaluated on the validation set that stays static and the evaluation score of the best is chosen at the end of each stage and reported as the accuracy of that stage including Stage 0.


Our model takes around 10 hours to train all 10 stages on a single Nvidia Titan Xp (12GB) GPU.


\noindent\textbf{Description of the Training and Sampling Procedure.}
The detailed description of the algorithms for training our model and active sampling are shown in \Aref{alg:training} and \Aref{alg:sampling} respectively.

\subsection{Additional Analysis}

\noindent\textbf{Other Approaches of Active Learning.}
As the comparison target of our approach, we also test different measures for distribution discrepancy.
In the following paragraphs, we introduce several possible candidates to measure the distances between single-modal and multi-modal answer representations.

\textbf{\emph{(1) KL-Divergence (KLD)}} is defined as follows:
\begin{equation}
    D_{KL}(\mathbf{y}||\mathbf{y}_q) = \sum_{i=1}^{|\mathcal{A}|}{\mathbf{y}^i\log(\mathbf{y}^i/\mathbf{y}^i_q)}, 
\end{equation} 
\begin{equation}
    D_{KL}(\mathbf{y}||\mathbf{y}_v) = \sum_{i=1}^{|\mathcal{A}|}{\mathbf{y}^i\log(\mathbf{y}^i/\mathbf{y}^i_v)}. 
\end{equation} 
Similar to our approach, by combining the two distance metrics via weighted sum, we have the KL-divergences:
\begin{equation}
    S(V,Q) = \alpha D_{KL}(\mathbf{y}||\mathbf{y}_q) + (1-\alpha) D_{KL}(\mathbf{y}||\mathbf{y}_v),
    \label{eqn:kld}
\end{equation}
where $\alpha$ is a hyper-parameter that weighs between visual and question scores.
Note that $S(V,Q)\geq0$, because KL-divergence is always non-negative.

\textbf{\emph{(2) Absolute Difference of KL-Divergence (AD-KLD)}} can also be used:
\begin{equation}
    S(V,Q) = |D_{KL}(\mathbf{y}||\mathbf{y}_q) - D_{KL}(\mathbf{y}||\mathbf{y}_v)|.
    \label{eqn:ad-kld}
\end{equation}
Although these methods have strong theoretical backing, we empirically show that for our problem, our proposed method shows the best performance.

\noindent\textbf{Ablation study for other approaches.}
Here, we introduce variants of our sampling methods.
\textbf{{\emph{KL-Divergence (KLD)}}} is described in \Eref{eqn:kld}. 
\textbf{\emph{Jensen-Shannon Divergence only (JSD)}} is the baseline that is sampled by only using Jensen-Shannon divergence described as: 
\begin{equation}
    JSD(\mathbf{y}_v||\mathbf{y}_q) = (D_{KL}(\mathbf{y}_v||M)+D_{KL}(\mathbf{y}_q||M))/2, 
\end{equation}
where  $M = (\mathbf{y}_v+\mathbf{y}_q)/2$. 
\textbf{\emph{Absolute Difference of KL-Divergence (AD-KLD)}} is described in \Eref{eqn:ad-kld}. 
\textbf{\emph{Mutual Information (MI)}} is the baseline that is sampled by only using mutual information: $S(V,Q) = \alpha I(A;V|Q) + (1-\alpha) I(A;Q|V)$. 
\textbf{\emph{Single-Modal Entropic Measure (SMEM) or (Ours)}} is our proposed method:
\begin{equation}
    S(V,Q) = \alpha H(Y_q) + (1-\alpha) H(Y_v) + \beta JSD(\mathbf{y}_v||\mathbf{y}_q) + \gamma H(\hat{Y}).
\end{equation}

The ablation study results are illustrated in \Fref{fig:ablation}.
We compare the AL performance of our model with other design choices for distribution discrepancy measures.
We draw the training curves for three different setups: (a) without distillation or entropy, (b) adding distillation loss only, and (c) adding distillation loss and entropy.
Each setup shows its strengths in different areas of the AL stages.
In (a), ours shows the best performance in the early stages with less data while \textbf{\emph{KLD}} shows better performance in the later stages as more data is accumulated.
In (b), ours shows the best performance with less data while \textbf{\emph{MI}} shows better performance as more data is accumulated.
In (c), we show that by taking entropy into account, our model shows the best performance compared to all the other design choices throughout all given stages.

\subsection{Qualitative Results}
\begin{figure*}[t]
\vspace{-2mm}
\centering
   \includegraphics[width=.7\linewidth]{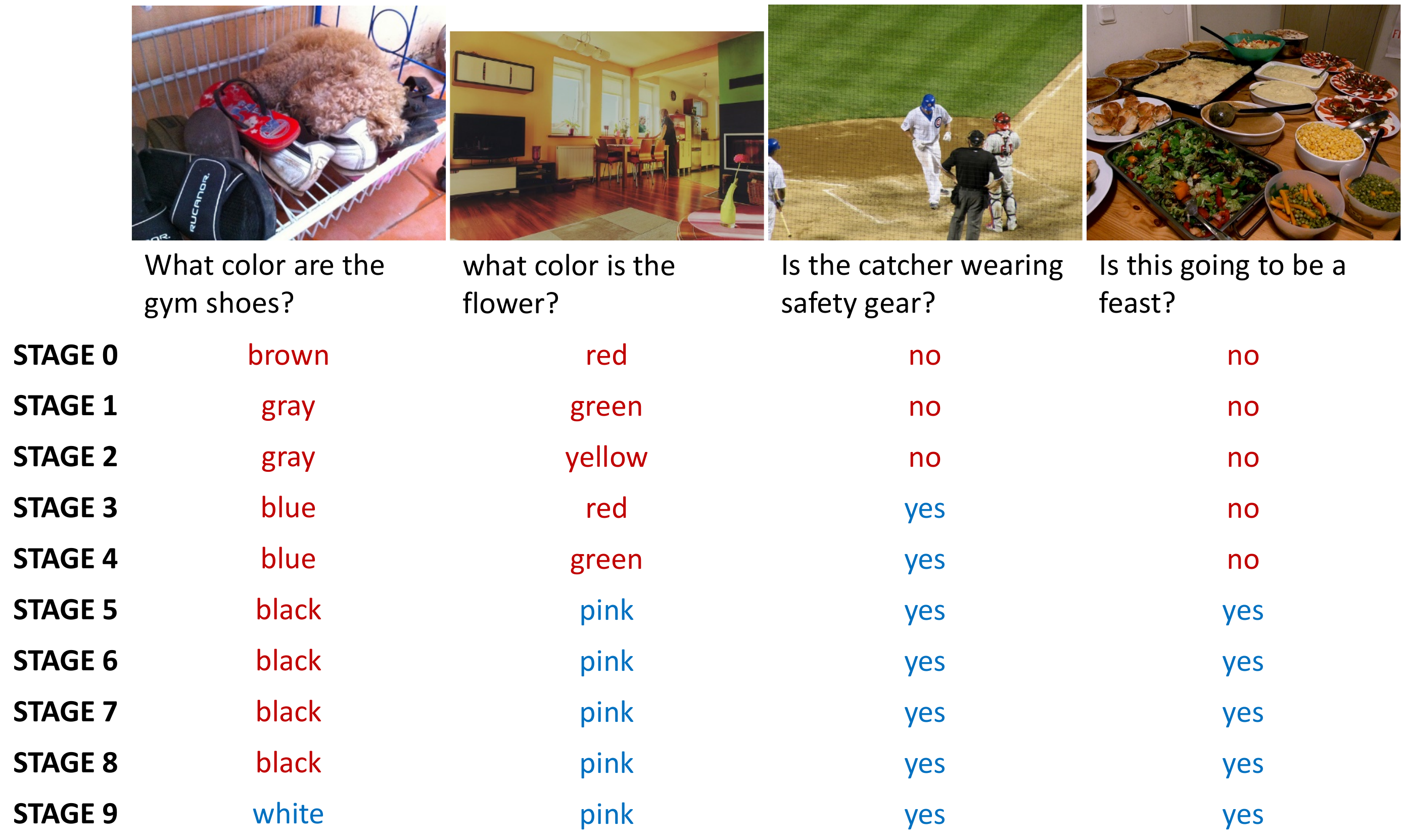}\\
   \includegraphics[width=.7\linewidth]{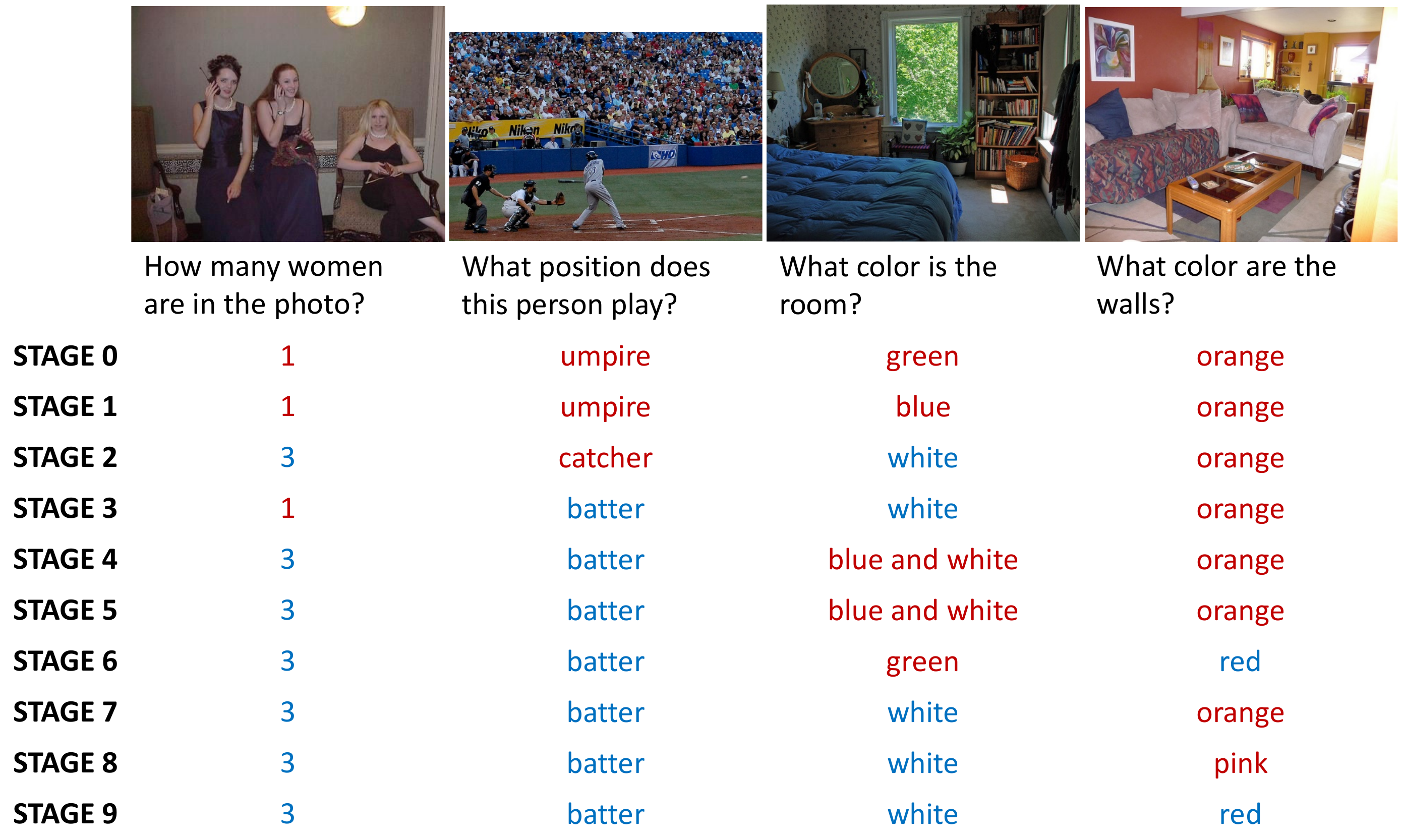}\\
   \includegraphics[width=.7\linewidth]{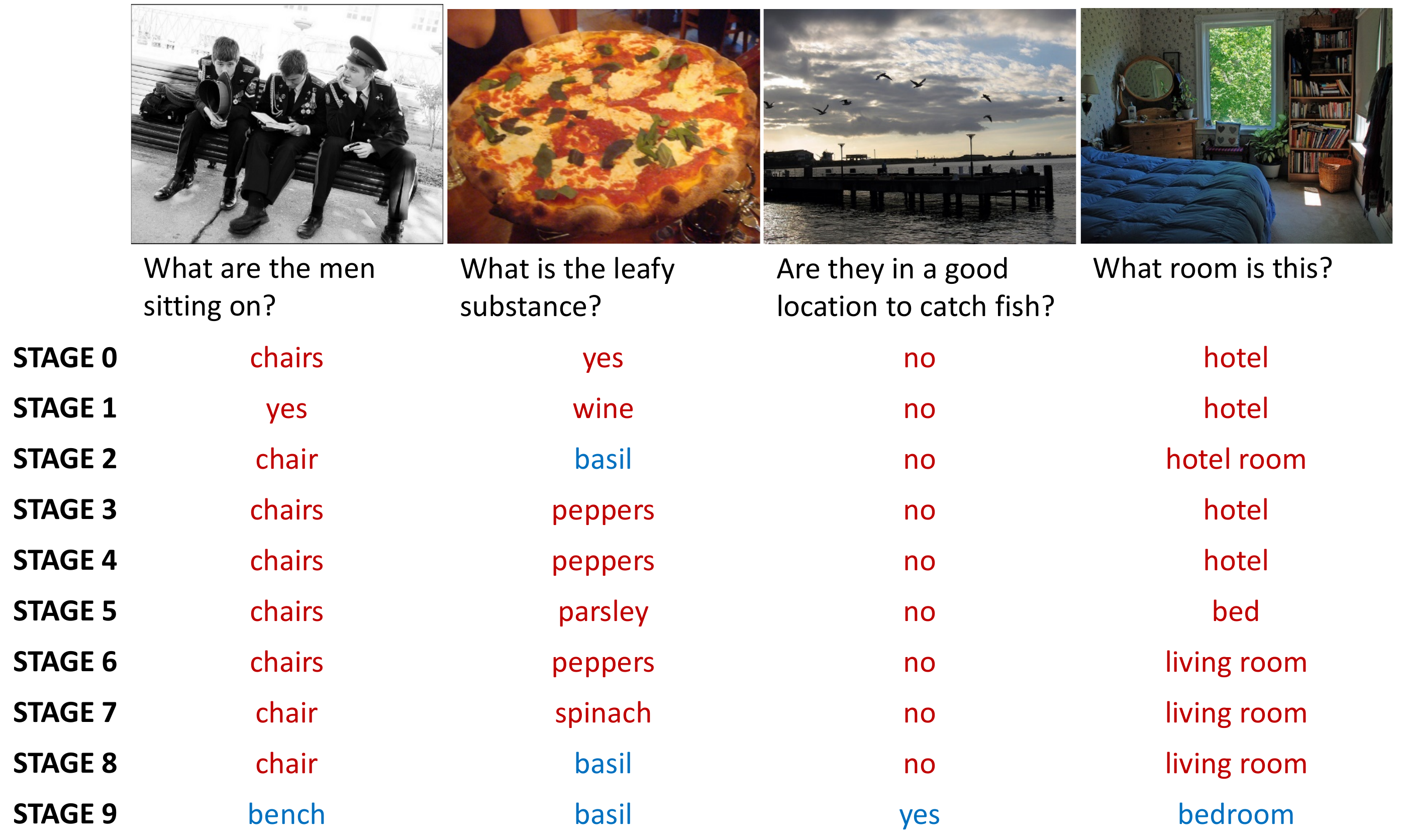}
   \caption{Qualitative result of the Top-1 prediction of our model for every Active Learning stage.}
\label{fig:qualitative}
\end{figure*}

\Fref{fig:qualitative} shows the qualitative results of our model for every AL stage. 
For each stage, we show the Top-1 predicted answer colored in blue if correct and red if not correct.
As the AL stages progress, the model is trained to generate increasingly accurate answers.

{\small
\bibliographystyle{ieee_fullname}
\bibliography{egbib}
}
